\pdfoutput=1

\documentclass[11pt]{article}

\usepackage[]{acl}

\usepackage{times}
\usepackage{latexsym}

\usepackage[T1]{fontenc}

\usepackage[utf8]{inputenc}

\usepackage{microtype}

\usepackage{inconsolata}
\usepackage[inkscapelatex=false]{svg}
\usepackage{latexsym}
\usepackage{multirow}
\usepackage{booktabs}
\usepackage[T1]{fontenc}
\usepackage{soul}
\usepackage{CJKutf8} 
\usepackage{pdflscape}
\usepackage[utf8]{inputenc}
\usepackage{threeparttable}
\usepackage{tabularx}
\usepackage{microtype}
\usepackage[figuresright]{rotating}
\usepackage{inconsolata}
\usepackage{xcolor}
\usepackage{amssymb}
\usepackage{pifont}
\usepackage{listings}
\usepackage{hyperref}

\usepackage{longtable}
\title{CMB: A Comprehensive  Medical Benchmark in Chinese}

\author{
 Xidong Wang$^*$, 
 Guiming Hardy Chen$^*$, 
 Dingjie Song$^*$, 
 Zhiyi Zhang, \\
 \textbf{Zhihong Chen, Qingying Xiao, Feng Jiang, Jianquan Li,} \\
 \textbf{Xiang Wan}, \textbf{Benyou Wang}$^\dag$, \textbf{Haizhou Li} \\
 The Chinese University of Hong Kong, Shenzhen\\
 Shenzhen Research Institute of Big Data\\
 \texttt{wangbenyou@cuhk.edu.cn} \\
}

\begin{document}
\begin{CJK}{UTF8}{gkai}
\maketitle

\renewcommand{\thefootnote}{\fnsymbol{footnote}}
\footnotetext[1]{Equal Contribution.}
\footnotetext[2]{Corresponding author.}
\renewcommand{\thefootnote}{\arabic{footnote}}

\begin{abstract}
Large Language Models (LLMs) provide a possibility to make a great breakthrough in medicine. The establishment of a standardized medical benchmark becomes a fundamental cornerstone to measure progression. 
However, medical environments in different regions have their local characteristics, e.g., the ubiquity and significance of traditional Chinese medicine within China. Therefore, merely translating English-based medical evaluation may result in \textit{contextual incongruities} to a local region. 
To solve the issue, we propose a localized medical benchmark called CMB, a Comprehensive  Medical Benchmark in Chinese, designed and rooted entirely within the native Chinese linguistic and cultural framework. While traditional Chinese medicine is integral to this evaluation, it does not constitute its entirety. 
Using this benchmark, we have evaluated several prominent large-scale LLMs, including ChatGPT, GPT-4, dedicated Chinese LLMs, and LLMs specialized in the medical domain.  
We hope this benchmark provide first-hand experience in existing LLMs for medicine and also  facilitate the widespread adoption and enhancement of medical LLMs within China.
Our data and code are publicly available at \href{https://github.com/FreedomIntelligence/CMB}{https://github.com/FreedomIntelligence/CMB}.

%
\end{abstract}
\section{Introduction}

Over the past two centuries, medical advancements have substantially increased human life expectancy. Medicine's effectiveness often hinges on experience, with veteran physicians typically outperforming novices. In parallel, large language models like ChatGPT are shaped by their vast data experiences. This mutual reliance on experiential learning between physicians and LLMs suggests a promising frontier for LLMs in the medical domain.

\textbf{Medical evaluation is highly professional.}
Although the future of \textit{LLMs for medicine} is promising, their evaluation is a challenging topic. Deploying LLMs in hospitals raises significant ethical concerns that real-world feedback becomes difficult.
Existing works on LLMs tend to leverage subjective evaluation~\citep{zheng2023judging} where none of references is used during the assessment. However, the evaluation in medicine is much more professional than that of the general domain. For instance, assessing \textit{radiology}-related issues poses a challenge for the public, a senior professor in medicine, or even a \textit{general practitioner}. 
Subjective evaluation would be difficult to be scaled up since professional manual judging is expensive.


\textbf{Benchmark for medical knowledge.}
Another school of evaluation protocol is objective evaluation, where the expected output has a clear reference. Certain protocols emphasize natural language understanding tasks that are not knowledge-intensive, as seen in studies \citep{zhang-etal-2022-cblue,peng2019transfer}. In the era of Large Language Models (LLM), modern NLP evaluations underscore the significance of knowledge~\citep{huang2023c,MMLU}. In biomedicine, a typical example to probe knowledge is BioLAMA~\cite{sung2021can}; however, it is tailored to evaluate masked language models instead of auto-regressive ones. Another benchmark is MultiMedBench~\cite{tu2023towards}, covering question answer, report summarization, visual question answering, report generation, and medical image classification. Note that MultiMedBench is only in English.

\textbf{The necessity to localize medical benchmark.}
During economic globalization, a unified medical standard may overlook the unique medical needs and practices of different regions and ethnic groups, indicating the necessity to localize medical benchmarks. 
For example, in Asia, 
Traditional Chinese Medicine (TCM) not only offers profound insights and localized medical solutions in the prevention, treatment, and rehabilitation of diseases but also has formed a medical paradigm closely associated with regional, climatic, dietary, and lifestyle characteristics, over its long historical evolution. In TCM, a disease has two aspects: "bìng" and "zhèng". The former is often translated as "disease entity". The latter, and more important one, is usually translated as "pattern". For example, the disease entity of a common cold might present with a pattern of wind-cold in one person, and with the pattern of wind-heat in another\footnote{\url{https://en.wikipedia.org/wiki/Traditional_Chinese_medicine\#Six_Excesses}}.

Simultaneously, it poses significant challenges when applying the Western medical framework to a local environment, which needs cross-cultural communication and understanding. In terms of disease diagnosis, examination of the tongue and the pulse are among the principal diagnostic methods in TCM. For example, redness on the tip of the tongue might indicate heat in the Heart, while redness on the sides of the tongue might indicate heat in the Liver. For drugs, there are roughly 13,000 compounds used in China and over 100,000 TCM recipes recorded in the ancient literature. Therefore, we should adopt a \textit{native} medical benchmark instead of a \textit{translated} medical benchmark for a local environment. Note that the precise translation of medical terminologies necessitates both medical professions and the cultural context in the target language.

\textbf{CMB’s  Philosophy.}
The CMB dataset comprises two parts: \textbf{CMB-Exam}, featuring multiple-choice questions from qualification exams, and \textbf{CMB-Clin}, including complex clinical diagnostic questions derived from real case studies. The dataset spans 6 major categories and 28 subcategories, totaling 280,839 multiple-choice questions. For \textbf{CMB-Exam}, we selected 400 questions from each subcategory to create an evaluation set. Additionally, \textbf{CMB-Clin} is formed from 74 expert-curated medical record consultations, drawn from clinical diagnostic teaching materials. Each multiple-choice question in the dataset offers four to six options, with one or more correct answers. The clinical diagnostic questions are based on real, intricate cases, with correct answers determined by a consensus of teaching experts.

\textbf{Take-away messages from CMB.}
After benchmarking various LLMs in CMB, we get the following observations that might be insightful.
\textbf{I)} GPT-4 and recent open-sourced LLMs such as Qwen-72B-Chat and Yi-34B-Chat, have achieved an accuracy rate exceeding 60\%, surpassing the threshold required for obtaining license;
\textbf{II) } Accuracy exhibits significant disparities across professional levels and knowledge areas, notably between \textbf{traditional Chinese medicine} and Western medicine;
\textbf{III)} The effectiveness of the \textbf{CoT  and few-shot prompts} varies among models with different accuracy levels, especially presenting potential risks in knowledge-intensive tasks;
and \textbf{IV) }Results of automatic evaluation using GPT-4 highly agree with \textbf{expert evaluation} results.

\begin{table*}[htbp]\scriptsize
\centering
\vspace{-10mm}
\resizebox{1\textwidth}{!}{
\begin{threeparttable}
\centering
\setlength{\tabcolsep}{3pt}
\begin{tabular}{@{}p{2.0cm}p{12.2cm}p{1.0cm}p{1.2cm}cc@{}lllll}
\toprule 
\textbf{Category} & \textbf{Subcategory} & \textbf{\# Subject} & \textbf{\# Questions} \\ 
\midrule
Physician  (医师) & Resident Physician  (住院医师); Licensed Assistant Physician  (执业助理医师); Licensed Physician  (执业医师); Associate Professional Physician  (中级职称); Advanced Professional Physicians  (高级职称) & 81 & 124,926\\
\midrule
Nurse  (护理) & Practicing Nurse  (护士); Licensed Practical Nurse  (护师); Charge Nurse  (主管护师); Advanced Practice Nurse  (高级护师) & 8 & 16,919\\
\midrule
Technicians  (医技) & Medical Technician  (医技士); Medical Technologist  (医技师); Supervising Technologist  (主管技师) & 21 & 27,004\\
\midrule
Pharmacist  (药师) & Licensed Pharmacist  (执业西药师); Licensed TCM Pharmacist  (执业中药师); Junior Pharmacist  (初级药师); Junior Pharmacist Assistant  (初级药士); Junior TCM Pharmacist  (初级中药师); Junior TCM Pharmacist Assistant  (初级中药士); Chief Pharmacists  (主管药师); Chief TCM Pharmacists  (主管中药师) & 8 & 33,354\\
\midrule
Undergraduate Disciplines  (学科考试)\tnote{1} & Fundamental Medicine  (基础医学); Clinical Medicine  (临床医学); Traditional Chinese  (TCM) and Chinese Herbal Medicine  (中医学与中药学); Preventive Medicine and Public Health  (预防医学与公共卫生学) & 53 & 62,271\\
\midrule
Graduate Entrance Exam  (考研) & Integrated Western Medicine  (西医综合); Integrated TCM  (中医综合); Political Science  (政治); Nursing  (护理学) & 5 & 16,365\\
\bottomrule
Total  & 28  & 176 & 280,839 \\
\bottomrule
\end{tabular}
\begin{tablenotes}
  \item [1] We referenced the National Standard Subject Classification of the People's Republic of China, see {\url{https://xkb.pku.edu.cn/docs/2018-10/20220328083301969071.pdf}}.
\end{tablenotes}
\end{threeparttable}
}
\caption{Statistics of the CMB-Exam Categories, Subcategories, Subjects, and Questions.}
\vspace{-2mm}
\label{tab:directories}
\end{table*}

\begin{table}[htbp]
\centering
\small
\begin{threeparttable}
\begin{tabularx}{0.5\textwidth}{lccc}
\toprule
Split &  \#subcategory  & \#Q  per subcategory &  \#Q in total  \\
\midrule
Test  & 28 & 400 & 11,200  \\
Dev& 28  & 10 \tnote{1} & 280 \\
Train & 28  & -\tnote{2} & 269,359 \\
\bottomrule
\end{tabularx}
\begin{tablenotes}
  \item [1] It is with explanations in dev set.
  \item [2] Each subcategory has a different number of questions.
\end{tablenotes}
\end{threeparttable}
\caption{Data split in CMB-Exam.}
\vspace{-2mm}
\end{table}

\section{The Philosophy of CMB}

\subsection{The Overall Philosophy}

We surveyed different medical professionals (physicians, nurses, technicians, and pharmacists) about the exams they encountered in their career development. Our research focused on common assessment types, leading us to select two key tasks for further study: multiple-choice questions and iterative questioning based on complex medical records. The former evaluates the model's knowledge grasp, while the latter assesses its practical problem-solving skills. Both tasks, having standard answers, provide reliable and stable performance indicators.


\subsection{Philosophy of CMB-Exam}

Existing medical benchmarks, sourced from the internet~\cite{li2023huatuo}, hospitals, etc., face privacy and accuracy challenges. We opted for qualification exams as our data source, creating the \textbf{CMB-Exam} subset. This choice is due to two key advantages: (I) qualification exams offer objective and typically accurate ground truths; (II) they provide a clear benchmark, namely a 60\% accuracy rate, which corresponds to the expertise level in specific domains. The multiple-choice questions in \textbf{CMB-Exam} encompass four clinical medical professions: \textit{physicians}, \textit{nurses}, \textit{medical technicians}, and \textit{pharmacists}. These exams span the entire professional journey, from undergraduate basics, graduate selections, standardized tests, professional qualifications, to intermediate and advanced professional title exams.

In the Chinese medical field, significant work has been done on multiple-choice tasks. MLEC-QA~\cite{li-etal-2021-mlec} compiled 21,700 manually annotated questions from the Chinese National Licensed Pharmacist Examination. 
Similarly, CMExam~\cite{CMExam} gathered 68,119 tagged questions from the same exam. 
However, it's important to note that the potential assistance provided by LLMs in medical professions is not confined to just pharmacy. It also encompasses a broad range of other health-related occupations, such as nursing and medical technology, among others.
Given that the Licensed Pharmacist Examination represents only a fraction of the career growth spectrum, its limited knowledge scope and occupational coverage do not provide detailed feedback. 
To address this, we compiled \textbf{CMB-Exam}, encompassing all medical-related occupations and the full range of exams encountered throughout their professional development.

\subsection{Philosophy of CMB-Clin}

Besides the theoretical exam content in \textbf{CMB-Exam}, the second subset, \textbf{CMB-Clin}, focuses on practical skills. This subset comprises complex clinical diagnostic problems to test the model's synthesis of knowledge and reasoning. It requires the model to utilize its medical knowledge for answering questions and to analyze case reports for informed responses. \textbf{CMB-Exam} and \textbf{CMB-Clin} together offer a comprehensive evaluation framework, applicable to both the career development of medical professionals and the learning trajectory of medical LLMs. To our knowledge, \textbf{CMB-Clin} is the inaugural multi-round question-answering dataset based on real, complex medical records.

\section{Dataset Creation}

\subsection{Taxonomy of CMB-Exam}
To obtain a precise taxonomy of medical evaluation, we aligned it with the disciplinary and examination systems of the medical field. 
First, we chose four main medical professions: physicians, pharmacists, medical technicians, and nurses, covering various occupational difficulty levels of examinations.
Considering the learning trajectories and professional growth paths, we additionally include \textit{discipline examinations} and \textit{graduate entrance examinations} for these four professions, ultimately resulting in six categories: Physician, Nurse, Technician, Pharmacist, Undergraduate Disciplines, and Graduate Entrance Exam.
One could refer to Table~\ref{tab:directories} for the detailed taxonomy. Moreover, we carried out a more detailed subject division within each subcategory, resulting in a total of 174 categories, the detailed directory list of which can be found in Appendix \ref{sec:appendix.catalog}. Through this structured arrangement, our directory structure reflects characteristics closely connected to the actual medical field, providing a solid foundation for further analysis and research.
\subsection{Creation of CMB-Exam}

\paragraph{Data Sources}
The data is derived from publicly available examination questions and coursework exercises with clear solutions provided by experts. A significant portion of these materials comes from the Medtiku\footnote{\url{https://www.medtiku.com/}}, from which we obtain explicit permission to share the data. 



\begin{table*}
    \vspace{-7mm}
    \centering
    \resizebox{1\textwidth}{!}{%
    \begin{tabular}{l|>{\centering\arraybackslash}p{1.5cm}>{\centering\arraybackslash}p{1.2cm}>{\centering\arraybackslash}p{1.2cm}>{\centering\arraybackslash}p{1.5cm}>{\centering\arraybackslash}p{2cm}>{\centering\arraybackslash}p{2.2cm}>{\centering\arraybackslash}p{1.5cm}>{\centering\arraybackslash}p{1.7cm}>{\centering\arraybackslash}p{2cm}>{\centering\arraybackslash}p{2cm}}
    \toprule
         \textbf{Department}&  \textbf{Internal Medicine}&  \textbf{Surgery}&  \textbf{Urology}&  \textbf{Neurology}&  \textbf{Hepatobiliary}&  \textbf{Endocrinology}&  \textbf{Pediatrics}&  \textbf{Gynecology}&  \textbf{Orthopedics}&\textbf{Dermatology} \\
    \midrule
         \textbf{Percentage (\%)}&  15.57&  14.87&  13.51&  12.16&  10.81&  8.11&  8.11&  6.76&  4.06&4.06\\
    \bottomrule
    \end{tabular}
    }
    \caption{Distribution of medical records across various departments.}
    \label{tab:clin_dist1}
\end{table*}

\begin{table*}
    \centering
    \resizebox{1\textwidth}{!}{%
    \begin{tabular}{l|>{\centering\arraybackslash}p{2cm}>{\centering\arraybackslash}p{1.5cm}>{\centering\arraybackslash}p{2cm}>{\centering\arraybackslash}p{2.8cm}>{\centering\arraybackslash}p{2.5cm}>{\centering\arraybackslash}p{2cm}>{\centering\arraybackslash}p{2cm}}
    \toprule
         \textbf{Diagnosis process}&  \textbf{Treatment principles}&  \textbf{Diagnosis}&  \textbf{Differential diagnosis}&  \textbf{Medical test recommendation}&  \textbf{Medical history analysis}&  \textbf{Operational knowledge}&  \textbf{Indications for surgery}\\
    \midrule
         \textbf{Percentage (\%)}&  30.09&  18.44&  14.32&  12.71&  11.65&  9.7&  3.09\\
    \bottomrule
    \end{tabular}
    }
    \caption{Distribution of questions located within the consultation process.}
    \vspace{-2mm}
    \label{tab:clin_dist2}
\end{table*}

\paragraph{Data Preprocessing}
Questions undergo a standardized data preprocessing procedure, including de-duplication and cleansing. For character recognition errors caused by OCR, we conduct a large amount of manual calibration to ensure the consistency with the original document. For possible grammatical or transcription errors, we use the comment system of Medtiku to delete data items with "problematic", "wrong question" and "incorrect" comments. Comparison of different directories before and after deletion is shown in Table~\ref{tab:Preprocess}.

\begin{table*}[]
\vspace{-7mm}
\resizebox{1\textwidth}{!}{
\begin{tabular}{@{}l|cccccc@{}}
\toprule
Category          & Physician & Nurse  & Technicians & Pharmacist & Undergraduate Disciplines & Graduate Entrance Exam \\ \midrule
Before Processing & 125,102   & 16,923 & 27,008      & 33,362     & 62,435                    & 16,367                 \\
After Processing  & 124,926   & 16,919 & 27,004      & 33,354     & 62,271                    & 16,365                 \\ \bottomrule
\end{tabular}}
\caption{Sample Numbers of different directories before and after Preprocessing}
\label{tab:Preprocess}
\end{table*}

\paragraph{Data Statistics}
Finally, we obtained a total of 280,839 multiple-choice questions.
To assess the model's comprehension of medical knowledge, we randomly selected 400 questions from each subcategory as a test set.
Additionally, to facilitate experiments with few-shot learning strategies, we randomly selected 10 questions from each subcategory as a development set.
We then enlisted the help of three medical specialists to generate explanations for each of these questions, specifically for the purpose of conducting chain-of-thought experiments (example shown in Figure~\ref{fig:CMB-Exam-cot-example}).
The remaining 269,359 questions were used as the train set.

\begin{figure*}[htbp]
\includegraphics[width=0.85\linewidth]{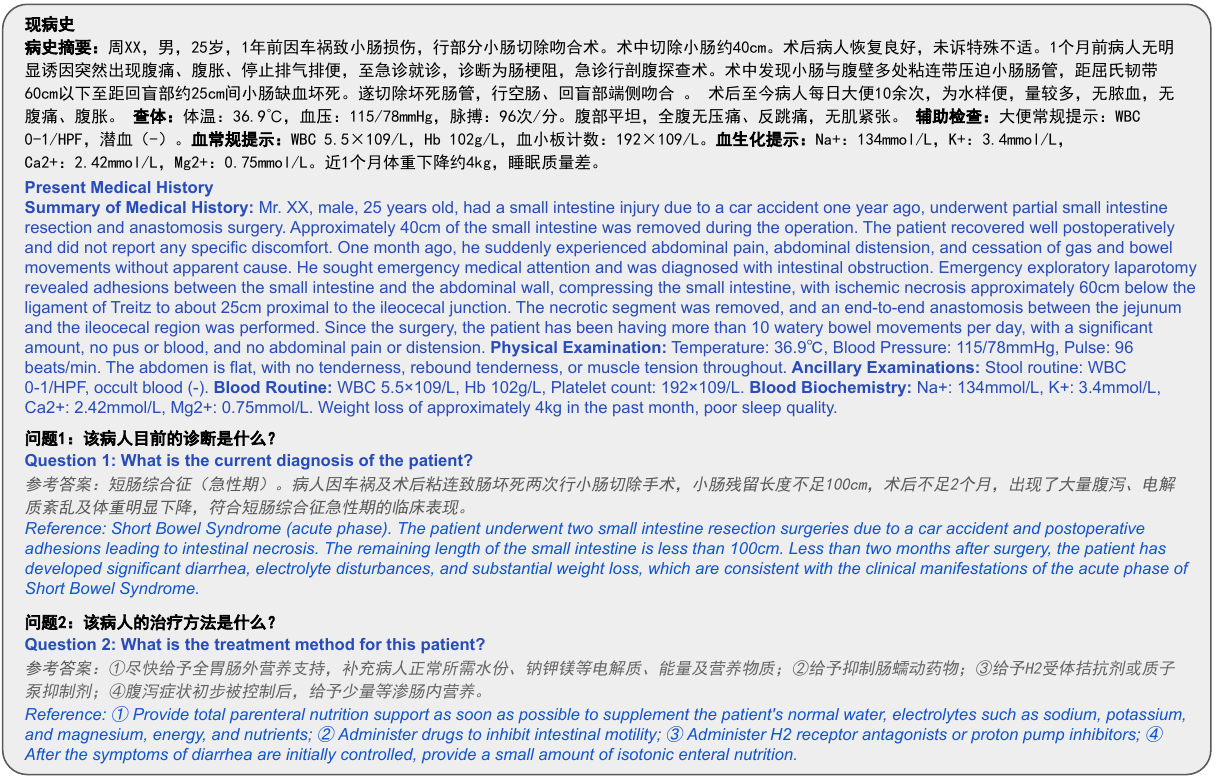}
\centering
\caption{Case of Short Bowel Syndrome from CMB-Clin. English translations are shown for better readability.}
\vspace{-2mm}
\label{fig:CMB-Clin-example}
\end{figure*}

\subsection{Creation of CMB-Clin}


CMB-Clin is designed to investigate models' proficiency in knowledge application amidst real-life diagnosis and treatment circumstances.

\paragraph{Data Preprocessing}
In order to obtain a high-quality dataset, we initially collected 108 cases with questions and answers from a variety of official medical textbooks.
These problems covered a wide range of disease types.
Subsequently, we performed quality screening based on the following criteria: 
Firstly, we eliminated problems that required reliance on image information, such as those that needed CT, MRI, and other imaging data for resolution. 
Secondly, we selected problems that contain sufficient diagnostic information to answer the questions.
Lastly, we removed cases with similar diseases to ensure the diversity of the problems. 
After such screening, we finally obtained 74 high-quality, complex, and real cases with 208 questions, as exemplified in Figure~\ref{fig:CMB-Clin-example}, for the construction of the CMB-Clin subset.

\paragraph{Task Formulation}

We transform the question answering task into the multi-turn dialogue task.
Specifically, for each case with questions, we simulate dialogue interactions between an \textbf{examiner} and a \textbf{candidate}, focusing on assessing the model's diagnostic and therapeutic capabilities.

\begin{table*}[ht]
\vspace{-2mm}
\scriptsize
\centering
\vspace{-5pt}
\resizebox{\textwidth}{!}{
\begin{tabular}{@{}ccccccccc@{}}
\toprule
\multicolumn{1}{l|}{\textbf{Model}}& \multicolumn{1}{c|}{\textbf{Open}} & \textbf{Physician}& \textbf{Nurse}& \textbf{Pharmacist}& \textbf{Technician}& \textbf{Disciplines}  & \multicolumn{1}{c|}{\textbf{Graduate Entrance Exam}} & \textbf{Average} \\ \midrule
\multicolumn{9}{c}{\textit{\textbf{General Models}}}\\ \midrule
\multicolumn{1}{l|}{Qwen-72B-Chat} & \multicolumn{1}{c|}{\multirow{2}{*}{$\checkmark$}} & \textbf{78.55 (80.00)} & \textbf{83.56 (84.06)} & 79.78 (\textbf{80.31}) & \textbf{77.92 (79.50)} & \textbf{68.26 (67.55)} & \multicolumn{1}{c|}{58.19 (\textbf{57.44})} & \textbf{74.38 (74.81)} \\
\multicolumn{1}{l|}{~+~CoT}& \multicolumn{1}{c|}{} & 78.40 (76.15) & 83.31 (81.69) & \textbf{80.13} (76.78) & 77.08 (73.00) & 67.69 (65.38) & \multicolumn{1}{c|}{\textbf{58.81} (55.50)} & 74.24 (71.42)\\ \midrule
\multicolumn{1}{l|}{Yi-34B-Chat} & \multicolumn{1}{c|}{\multirow{2}{*}{$\checkmark$}} & 71.10 (72.95) & 77.56 (80.44) & 73.16 (74.03) & 73.67 (76.92) & 66.56 (67.31) & \multicolumn{1}{c|}{52.94 (55.63)} & 69.17 (71.22) \\
\multicolumn{1}{l|}{~+~CoT}& \multicolumn{1}{c|}{} & 69.05 (58.45) & 74.75 (63.13) & 70.25 (60.06) & 68.00 (57.08) & 63.00 (56.31) & \multicolumn{1}{c|}{51.69 (45.44)} & 66.12 (56.75)\\ \midrule
\multicolumn{1}{l|}{Yi-6B-Chat} & \multicolumn{1}{c|}{\multirow{2}{*}{$\checkmark$}} & 67.25 (68.70) & 76.38 (77.06) & 68.50 (69.38) & 67.83 (68.00) & 61.75 (62.44) & \multicolumn{1}{c|}{53.50 (55.88)} &  65.87 (66.91) \\
\multicolumn{1}{l|}{~+~CoT}& \multicolumn{1}{c|}{} & 64.30 (59.95) & 73.50 (67.38) & 64.44 (61.03) & 65.33 (56.67) & 59.63 (55.25) & \multicolumn{1}{c|}{50.50 (48.75)} & 62.95 (58.17)\\ \midrule
\multicolumn{1}{l|}{GPT-4}  & \multicolumn{1}{c|}{\multirow{2}{*}{\ding{55}}}  & 59.90  (60.19) & 69.31  (70.14) & 52.19  (53.25) & 61.50  (62.38) & 59.69  (60.25) & \multicolumn{1}{c|}{54.19  (55.12)}  & 59.46  (60.22) \\ 
\multicolumn{1}{l|}{~+~CoT}& \multicolumn{1}{c|}{} & 58.15  (59.63) & 70.31  (71.02) & 53.09  (52.15) & 62.34  (61.38) & 60.69  (62.25) & \multicolumn{1}{c|}{52.30  (54.21)}  & 59.45  (60.11)  \\ \midrule 

\multicolumn{1}{l|}{Deepseek-llm-67B-Chat} & \multicolumn{1}{c|}{\multirow{2}{*}{$\checkmark$}} & 52.90 (59.15) & 61.50 (65.19) & 54.28 (59.22) & 51.42 (55.25) & 51.19 (55.63) & \multicolumn{1}{c|}{40.63 (38.88)} & 51.99 (55.55) \\
\multicolumn{1}{l|}{~+~CoT}& \multicolumn{1}{c|}{} & 56.20 (51.80) & 60.19 (60.25) & 54.44 (53.69) & 50.75 (47.58) & 51.38 (51.63) & \multicolumn{1}{c|}{41.00 (38.50)} & 52.33 (50.58) \\ \midrule
\multicolumn{1}{l|}{Baichuan2-13B-Chat} & \multicolumn{1}{c|}{\multirow{2}{*}{$\checkmark$}} & 49.55 (50.05) & 56.75 (57.75) & 49.41 (50.50) & 50.08 (49.50) & 48.25 (49.06) & \multicolumn{1}{c|}{39.18 (40.63)} & 48.87 (49.58) \\
\multicolumn{1}{l|}{~+~CoT}& \multicolumn{1}{c|}{} & 48.90 (48.55) & 57.25 (54.75) & 49.88 (49.16) & 49.33 (47.08) & 46.88 (44.37) & \multicolumn{1}{c|}{38.06 (36.44)} & 48.38 (46.73) \\ \midrule
\multicolumn{1}{l|}{Qwen-7B-Chat} & \multicolumn{1}{c|}{\multirow{2}{*}{$\checkmark$}} & 48.00 (49.45) & 54.25 (55.13) & 48.34 (47.94) & 48.08 (49.25) & 44.87 (45.00) & \multicolumn{1}{c|}{35.94 (36.56)} & 46.58 (47.22) \\
\multicolumn{1}{l|}{~+~CoT}& \multicolumn{1}{c|}{} & 48.00 (45.65) & 54.25 (52.63) & 48.34 (47.28) & 48.08 (43.08) & 44.88 (44.06) & \multicolumn{1}{c|}{35.94 (36.19)} & 46.58 (44.82)\\ \midrule
\multicolumn{1}{l|}{Baichuan2-7B-Chat} & \multicolumn{1}{c|}{\multirow{2}{*}{$\checkmark$}} & 42.55 (43.30) & 51.75 (51.56) & 44.59 (44.59) & 45.50 (43.00) & 43.00 (40.44) & \multicolumn{1}{c|}{32.56 (34.00)} & 43.33 (42.82) \\
\multicolumn{1}{l|}{~+~CoT}& \multicolumn{1}{c|}{} & 43.55 (38.25) & 51.06 (47.13) & 44.72 (43.91) & 43.17 (39.50) & 42.69 (39.63) & \multicolumn{1}{c|}{32.44 (20.56)} & 42.94 (38.16)\\ \midrule
\multicolumn{1}{l|}{ChatGLM3-6B} & \multicolumn{1}{c|}{\multirow{2}{*}{$\checkmark$}} & 42.55 (40.30) & 47.31 (44.81) & 39.56 (38.22) & 41.08 (39.33) & 37.44 (37.63) & \multicolumn{1}{c|}{32.06 (32.13)} & 40.00 (38.74)  \\
\multicolumn{1}{l|}{~+~CoT}& \multicolumn{1}{c|}{} & 38.90 (38.15) & 45.38 (43.25) & 38.19 (34.53) & 38.08 (37.08) & 36.12 (35.25) & \multicolumn{1}{c|}{30.13 (26.75)} & 37.80 (35.84)\\ \midrule
\multicolumn{1}{l|}{ChatGPT}& \multicolumn{1}{c|}{\multirow{2}{*}{\ding{55}}} & 40.75  (40.75)  & 45.69  (45.69)  & 36.59  (36.59)  & 40.08  (40.08)  & 37.94  (37.94)  & \multicolumn{1}{c|}{28.81  (28.81)}& 38.31  (38.31)  \\
\multicolumn{1}{l|}{~+~CoT}& \multicolumn{1}{c|}{}  & 17.75  (17.75)  & 19.94  (19.94)  & 16.00  (16.00)  & 20.25  (20.25)  & 19.25  (19.25)  & \multicolumn{1}{c|}{16.19  (16.19)}& 18.23  (18  .23) \\ \midrule
\multicolumn{1}{l|}{Internlm-Chat-20B} & \multicolumn{1}{c|}{\multirow{2}{*}{$\checkmark$}} & 39.35 (39.55) & 45.44 (43.00) & 38.53 (36.25) & 37.92 (38.25) & 38.12 (38.06) & \multicolumn{1}{c|}{29.63 (29.63)} & 38.17 (37.46) \\
\multicolumn{1}{l|}{~+~CoT}& \multicolumn{1}{c|}{} & 39.60 (34.00) & 44.44 (41.00) & 36.41 (32.50) & 40.08 (34.17) & 37.88 (32.81) & \multicolumn{1}{c|}{30.88 (26.19)} & 38.21 (33.45)\\ \midrule
\multicolumn{1}{l|}{Internlm-Chat-7B} & \multicolumn{1}{c|}{\multirow{2}{*}{$\checkmark$}} & 34.45 (32.55) & 42.13 (36.81) & 33.69 (32.41) & 37.50 (35.00) & 33.75 (31.06) & \multicolumn{1}{c|}{27.94 (26.94)} & 34.91 (32.46) \\
\multicolumn{1}{l|}{~+~CoT}& \multicolumn{1}{c|}{} & 35.55 (34.70) & 41.38 (38.31) & 33.88 (32.41) & 35.83 (35.42) & 33.88 (32.19) & \multicolumn{1}{c|}{27.88 (25.50)} & 34.73 (33.09) \\ \midrule

\multicolumn{1}{l|}{Mixtral-8x7B-32kseqlen} & \multicolumn{1}{c|}{\multirow{2}{*}{$\checkmark$}} & 34.50 (39.00) & 42.00 (41.88) & 25.06 (33.13) & 30.83 (39.50) & 31.81 (36.44) & \multicolumn{1}{c|}{22.25 (28.25)} &  31.07 (36.37)\\ 
\multicolumn{1}{l|}{~+~CoT}& \multicolumn{1}{c|}{} & 34.50 (28.00) & 42.00 (34.06) & 25.06 (24.69) & 30.83 (34.75) & 31.81 (27.50) & \multicolumn{1}{c|}{22.25 (17.56)} & 31.07 (27.76)\\  \midrule
 

\multicolumn{1}{l|}{Qwen-1.8B-Chat} & \multicolumn{1}{c|}{\multirow{2}{*}{$\checkmark$}} & 26.20 (44.15) & 34.06 (50.63) & 28.03 (39.78) & 27.08 (39.25) & 25.69 (36.56) & \multicolumn{1}{c|}{23.50 (33.75)} & 27.43 (40.69) \\
\multicolumn{1}{l|}{~+~CoT}& \multicolumn{1}{c|}{}  & 26.20 (30.95) & 34.06 (41.50) & 28.03 (32.25) & 27.08 (28.00) & 25.69 (27.81) & \multicolumn{1}{c|}{23.50 (28.00)} & 27.43 (31.42)\\ \midrule

\multicolumn{1}{l|}{Mistral-7B-Instruct-v0.1} & \multicolumn{1}{c|}{\multirow{2}{*}{$\checkmark$}} & 23.75 (19.55) & 22.19 (22.50) & 20.97 (19.88) & 25.83 (21.42) & 21.94 (19.25) & \multicolumn{1}{c|}{18.88 (16.75)} &  22.26 (19.89)\\
\multicolumn{1}{l|}{~+~CoT}& \multicolumn{1}{c|}{} & 21.90 (19.95) & 23.06 (21.44) & 20.97 (19.97) & 23.08 (20.83) & 21.81 (19.00) & \multicolumn{1}{c|}{15.56 (12.94)} & 19.02 (19.02)\\ 
\bottomrule
\end{tabular}
}
\caption{Accuracy in the answer-only and CoT settings across different categories for \textbf{general models}. Values in parentheses are the Three-shot accuracy.}

\label{table:general}
\end{table*}

Figure~\ref{fig:CMB-Clin-example} illustrates the structure of each case, which includes three main parts:
\textbf{I) Description} $D$: patient information, including medical history summaries and chief complaints, physical examinations  (e.g., visual and tactile inspection), and ancillary examinations  (e.g., biopsy and CT scans); 
\textbf{II)  Questions} $Q$: questions related to diagnosis and treatment based on descriptions. Some questions might be interrelated;
and  \textbf{III) References} $R$: corresponding reference answers to questions.

Formally, to simulate dialogue interactions, we concatenate the patient's description with the previous question-answer pairs and the current question in each conversation round  (e.g., the \(k\)-th round). 
This concatenated input, denoted as \(x\), is represented as follows:
$x = D_i + Q_i + R_i + \ldots Q_{i+k}$.
The reference answer for this input is \(R_{i+k}\).
For the response \(\hat{R}_{i+k}\), we will evaluate it from four dimensions, including: Fluency, Relevance, Completeness, and Proficiency. These dimensions are adopted as per suggestions from experts.
%

\paragraph{Data Distribution}
We commissioned a medical professional to examine the distribution of questions within the CMB-Clin from two perspectives: the department associated with the medical record, and the point in the consultation process where the question arises.
Table~\ref{tab:clin_dist1} displays the distribution of medical records across various departments. It reveals a wide-ranging coverage, with internal medicine and surgery representing the most substantial segments (15.57\% and 14.87\% respectively).
Table~\ref{tab:clin_dist2} illustrates where in the consultation process the questions are situated. It indicates that questions permeate the entire diagnostic procedure, with a significant portion arising during the treatment principles stage (30.09\%).

\section{Experiments on CMB-Exam}
\subsection{Experimental Setup}
\paragraph{Models} We evaluate the following Chinese medical LLMs to compare their performance on CMB-Exam: HuatuoGPT-II~\citep{chen2023huatuogptii}, ChatMed-Consult~\citep{zhu2023ChatMed}, MedicalGPT~\citep{MedicalGPT}, ChatGLM-Med~\citep{ChatGLM-Med}, DoctorGLM~\cite{xiong2023doctorglm}, BianQue-2~\cite{chen2023bianque1}, Bentsao~\citep{wang2023huatuo}, IvyGPT~\citep{wang2023ivygpt}, Sunsimiao~\citep{Sunsimiao} and DISC-MedLLM~\citep{bao2023discmedllm}. In addition to these specialized models, we also evaluate some publicly-available general-domain instruction-following model series, namely ChatGLM3-6B~\citep{du2022glm}, Baichuan2~\cite{baichuan2023baichuan2}, Qwen~\cite{qwen}, Yi\footnote{\url{https://github.com/01-ai/Yi}}, Deepseek-llm~\citep{deepseek-llm}, Mistral~\cite{jiang2023mistral} and Internlm~\cite{2023internlm}. For closed source commercial models, we evaluate ChatGPT\footnote{We use the version \texttt{gpt-3.5-turbo-16k-0613}.}, GPT-4\footnote{We use the version \texttt{gpt-4}.}, ShunkunGPT, AntGLM-Med and JianpeiGPT. We also test the performance of DISC-MedLLM trained on CMB-Exam-Train. All experiments are conducted in August, 2023. Please refer to Appendix \ref{sec:appendix.model} for more details about models and training.

\paragraph{Decoding Hyperparameters}

For all the aforementioned open source models, we adopt their default hyper-parameters. Besides, to reduce the variance in generation, we adopt greedy decoding for all models on both CMB-Exam and CMB-Clin.
And the \texttt{min\_new\_tokens} and \texttt{max\_new\_tokens} are set to 1 and 512, respectively, to avoid empty or lengthy answers on CMB-Exam.

\begin{table*}[ht]
\scriptsize
\centering
\vspace{-10pt}
\resizebox{\textwidth}{!}{
\begin{tabular}{@{}ccccccccc@{}}
\toprule
\multicolumn{1}{l|}{\textbf{Model}}& \multicolumn{1}{c|}{\textbf{Open}} & \textbf{Physician}& \textbf{Nurse}& \textbf{Pharmacist}& \textbf{Technician}& \textbf{Disciplines}  & \multicolumn{1}{c|}{\textbf{Graduate Entrance Exam}} & \textbf{Average} \\ \midrule

\multicolumn{9}{c}{\textit{\textbf{Commercial Medical Models}}} \\
\midrule
\multicolumn{1}{l|}{JianPeiGPT}& \multicolumn{1}{c|}{\multirow{1}{*}{\ding{55}}} & 73.60* & 77.00* & 72.84* & 65.00* & 70.13* & \multicolumn{1}{c|}{78.40*} & 72.84* \\ \midrule
\multicolumn{1}{l|}{ShuKunGPT}& \multicolumn{1}{c|}{\multirow{1}{*}{\ding{55}}} & 68.65* & 71.44* & 70.78* & 61.92* & 62.81* & \multicolumn{1}{c|}{51.06*} & 64.44* \\ \midrule
\multicolumn{1}{l|}{AntGLM-Med}& \multicolumn{1}{c|}{\multirow{1}{*}{\ding{55}}} & 62.85* & 66.81* & 60.06* & 48.50* & 54.69* & \multicolumn{1}{c|}{51.06*} & 55.00* \\ \midrule
\multicolumn{9}{c}{\textit{\textbf{Open source Medical Models}}} \\
\midrule
\multicolumn{1}{l|}{HuatuoGPT-II-34B  (华佗II)} & \multicolumn{1}{c|}{\multirow{2}{*}{$\checkmark$}} & 75.65 (75.65) & 82.31 (\textbf{82.31}) & \textbf{76.81 (77.12)} & 76.17 (\textbf{74.12}) & \textbf{74.38 (74.38)} & \multicolumn{1}{c|}{\textbf{75.56} (75.56)} & \textbf{76.82 (76.52)} \\
\multicolumn{1}{l|}{~+~CoT}& \multicolumn{1}{c|}{} & \textbf{76.13 (76.13)} & \textbf{83.15} (82.15) & \textbf{76.81} (76.81) & \textbf{77.12} (70.12) & 71.22 (70.22) & \multicolumn{1}{c|}{\textbf{75.56 (76.12)}} & 76.67 (75.26)\\ \midrule
\multicolumn{1}{l|}{HuatuoGPT-II-13B  (华佗)} & \multicolumn{1}{c|}{\multirow{2}{*}{$\checkmark$}} & 67.85 (67.85) & 66.12 (66.12) & 64.91 (64.91) & 62.00 (63.05) & 61.94 (62.15) & \multicolumn{1}{c|}{53.69 (54.69)} & 62.75 (63.13) \\
\multicolumn{1}{l|}{~+~CoT}& \multicolumn{1}{c|}{} & 68.02 (68.05) & 65.32 (65.32) & 65.12 (65.12) & 63.01 (62.55) & 62.01 (61.53) & \multicolumn{1}{c|}{54.60 (54.63)} & 63.01 (62.87)\\ \midrule
\multicolumn{1}{l|}{HuatuoGPT-II-7B  (华佗)} & \multicolumn{1}{c|}{\multirow{2}{*}{$\checkmark$}} & 64.55 (64.55) & 63.75 (63.75) & 61.06 (61.06) & 56.25 (56.25) & 56.63 (56.90) & \multicolumn{1}{c|}{51.81 (53.82)} & 59.00 (59.39) \\
\multicolumn{1}{l|}{~+~CoT}& \multicolumn{1}{c|}{} & 65.12 (65.12) & 64.33 (63.12) & 60.05 (61.50) & 57.12 (56.03) & 56.63 (57.01) & \multicolumn{1}{c|}{51.81 (52.18)} & 59.18 (59.16)\\ \midrule
\multicolumn{1}{l|}{DISC-MedLLM-13B} & \multicolumn{1}{c|}{\multirow{2}{*}{$\checkmark$}} & 42.25 (42.20) & 46.88 (47.87) & 38.44 (38.94) & 38.83 (38.92) & 40.75 (39.38) & \multicolumn{1}{c|}{31.44 (31.25)} &  39.76 (39.76)\\
\multicolumn{1}{l|}{~+~CoT}& \multicolumn{1}{c|}{} & 41.85 (41.30) & 47.19 (46.44) & 38.97 (38.41) & 39.17 (38.17) & 40.31 (39.81) & \multicolumn{1}{c|}{31.37 (31.44)} & 39.78 (39.26)\\ \midrule
\multicolumn{1}{l|}{IvyGPT-13B} & \multicolumn{1}{c|}{\multirow{2}{*}{$\checkmark$}} & 37.70 (37.34) & 43.56 (43.56) & 40.47 (41.25) & 38.08 (39.06) & 35.31 (36.31) & \multicolumn{1}{c|}{36.12 (37.15)} & 38.54 (39.11) \\
\multicolumn{1}{l|}{~+~CoT}& \multicolumn{1}{c|}{} & 37.15 (38.23) & 44.12 (45.12) & 41.23 (42.33) & 38.08 (39.12) & 36.12 (37.20) & \multicolumn{1}{c|}{36.12 (36.88)} & 38.80 (39.81)\\ \midrule
\multicolumn{1}{l|}{Sunsimiao-7B  (孙思邈)} & \multicolumn{1}{c|}{\multirow{2}{*}{$\checkmark$}} & 38.75 (38.12) & 44.37 (45.12) & 38.81 (39.12) & 38.33 (38.33) & 37.50 (38.12) & \multicolumn{1}{c|}{33.31 (34.21)} & 38.51 (33.13) \\
\multicolumn{1}{l|}{~+~CoT}& \multicolumn{1}{c|}{} & 39.12 (39.12) & 45.12 (45.12) & 38.81 (39.12) & 38.33 (39.31) & 37.50 (38.12) & \multicolumn{1}{c|}{34.12 (34.12)} & 38.84 (39.96)\\ \midrule


\multicolumn{1}{l|}{MedicalGPT-7B}& \multicolumn{1}{c|}{\multirow{2}{*}{$\checkmark$}} & 26.40  (26.56)  & 30.94  (30.94)  & 24.72  (24.84)  & 27.17  (27.32)  & 25.44  (25.62)  & \multicolumn{1}{c|}{21.50  (21.64)}& 26.03  (26.15)  \\
\multicolumn{1}{l|}{~+~CoT}& \multicolumn{1}{c|}{}  & 24.80  (25.61)  & 27.19  (27.98)  & 23.09  (24.07)  & 24.58  (26.00)  & 23.75  (24.77)  & \multicolumn{1}{c|}{21.06  (21.79)}& 24.08  (25.04)  \\ \midrule
\multicolumn{1}{l|}{ChatMed-Consult-7B}& \multicolumn{1}{c|}{\multirow{2}{*}{$\checkmark$}} & 20.20  (21.41)  & 22.31  (23.48)  & 20.59  (21.58)  & 22.67  (23.55)  & 20.38  (21.36)  & \multicolumn{1}{c|}{17.44  (18.08)}& 20.60  (21.58)  \\
\multicolumn{1}{l|}{~+~CoT}& \multicolumn{1}{c|}{}  & 19.40  (20.92)  & 21.69  (23.56)  & 20.00  (21.65)  & 22.83  (23.59)  & 18.88  (20.44)  & \multicolumn{1}{c|}{18.56  (19.55)}& 20.23  (21.62)  \\ \midrule
\multicolumn{1}{l|}{ChatGLM-Med-7B}& \multicolumn{1}{c|}{\multirow{2}{*}{$\checkmark$}} & 21.75  (23.59)  & 22.06  (23.37)  & 21.84  (22.67)  & 21.00  (21.85)  & 18.44  (19.72)  & \multicolumn{1}{c|}{17.50  (18.14)}& 20.43  (21.56)  \\
\multicolumn{1}{l|}{~+~CoT}& \multicolumn{1}{c|}{}  & 15.55  (20.89)  & 16.25  (22.13)  & 17.34  (21.06)  & 16.33  (20.65)  & 12.63  (17.12)  & \multicolumn{1}{c|}{12.56  (16.88)}& 15.11  (19.79)  \\ \midrule
\multicolumn{1}{l|}{Bentsao-7B  (本草)}& \multicolumn{1}{c|}{\multirow{2}{*}{$\checkmark$}} & 21.55  (21.67)  & 19.94  (19.99)  & 20.94  (21.07)  & 22.75  (22.85)  & 19.56  (19.83)  & \multicolumn{1}{c|}{16.81  (16.93)}& 20.26  (20.39)  \\
\multicolumn{1}{l|}{~+~CoT}& \multicolumn{1}{c|}{}  & 21.00  (21.10)  & 20.56  (20.61)  & 20.66  (20.78)  & 22.17  (22.24)  & 19.25  (19.53)  & \multicolumn{1}{c|}{16.44  (16.54)}& 20.01  (20.13)  \\ \midrule
\multicolumn{1}{l|}{BianQue-2 (扁鹊-2)} & \multicolumn{1}{c|}{\multirow{2}{*}{$\checkmark$}} & 4.90  (4.40)  & 4.19  (5.19)  & 4.28  (7.97)  & 3.58  (8.08)  & 3.31  (5.69)  & \multicolumn{1}{c|}{3.25  (4.00)}& 3.92  (5.89)  \\ 
\multicolumn{1}{l|}{~+~CoT}& \multicolumn{1}{c|}{} & 7.85 (6.95) & 6.63 (7.31) & 7.34 (7.25) & 8.33 (9.75) & 6.63 (6.94) & \multicolumn{1}{c|}{5.94 (6.06)} & 7.12 (7.38)\\ \midrule
\multicolumn{1}{l|}{DoctorGLM} & \multicolumn{1}{c|}{\multirow{2}{*}{$\checkmark$}} & 2.70  (0.10)  & 3.31  (0.38)  & 3.84  (0.34)  & 3.75  (0.50)  & 3.19  (0.37)  & \multicolumn{1}{c|}{2.25  (0.81)}& 3.17  (0.42)  \\ 
\multicolumn{1}{l|}{~+~CoT}& \multicolumn{1}{c|}{} & 3.15 (2.35) & 3.13 (2.50) & 3.41 (3.28) & 2.50 (1.17) & 3.38 (3.06) & \multicolumn{1}{c|}{2.25 (3.88)} & 2.97 (2.71)\\ \midrule
\multicolumn{9}{c}{\textit{\textbf{Models Trained by CMB-Exam-Train}}}\\ \midrule
\multicolumn{1}{l|}{DISC-MedLLM-13B  (CMB-Exam-Train)} & \multicolumn{1}{c|}{\multirow{2}{*}{$\checkmark$}} & 43.22 (43.22) & 48.13 (47.56) & 39.12 (40.23) & 40.12 (45.12) & 41.25 (42.25) & \multicolumn{1}{c|}{33.25 (33.75)} & 40.85 (42.02) \\
\multicolumn{1}{l|}{~+~CoT}& \multicolumn{1}{c|}{} & 42.65 (43.65) & 47.15 (48.13) & 40.12 (41.22) & 39.32 (40.12) & 42.25 (41.58) & \multicolumn{1}{c|}{33.80 (34.80)} & 40.88 (41.58)\\
\bottomrule

\end{tabular}
}
\caption{Accuracy in the answer-only and CoT settings across different categories for \textbf{medical models}. Values in parentheses are the Three-shot accuracy. * means we only have the best score and the generation strategy is unknown.}
\vspace{-2pt}
\label{table:medical}
\end{table*}

\paragraph{Evaluation Details}
We evaluate the models in both answer-only and chain-of-thought  (CoT) settings.
We extract answers from model outputs using an empirically designed regular expression. Each extracted answer is compared to the solution and is deemed correct if and only if they are exactly matched. We adopt accuracy as our metric.
All evaluation experiments and training experiments take around 1000 GPU-hours on 8 NVIDIA A800 80GB GPUs.

\subsection{Benchmarking Results}

We report the results in Table~\ref{table:general} and Table~\ref{table:medical}. 
There are several observations drawn from different aspects.

\begin{figure*}[ht]
\includegraphics[width=1\linewidth]{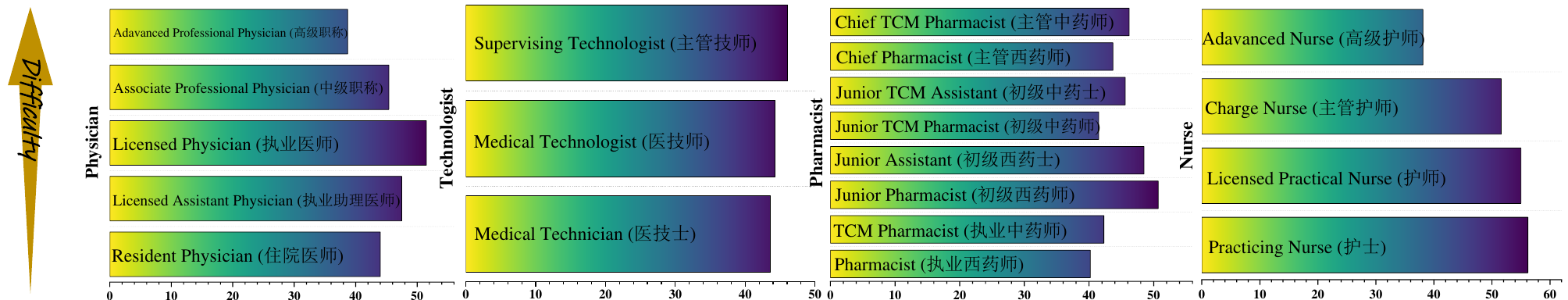}
\centering
\caption{Accuracy across various clinical medicine fields at different career stages. The accuracies are the Zero-shot average values for all the models using direct response strategy. The difficulty increases from bottom to top.}
\label{fig:difficulty}
\end{figure*}

\paragraph{On General LLMs.} As shown in Table~\ref{table:general}, there is no general model that is particularly lacking in medical knowledge. Taking GPT-4 and ChatGPT as the watershed, Qwen-72B, Yi-34B, and Yi-6B have achieved accuracy exceeding GPT-4, and most models have achieved performance exceeding ChatGPT. Yi-6B achieves such good results with a smaller number of parameters is impressive, but it also reminds us of the possibility of data leakage. At the same time, we also noticed that Qwen-1.8B showed strong in-context learning capabilities. Considering its number of parameters, this is also an exciting discovery. Domestic general models have completed catching up with OpenAI in medical knowledge in chinese.

\paragraph{On Medical LLMs.} As shown in Table 4, the gap between medical models is obvious. HuatuoGPT-II surpasses all commercial models and ranks first, demonstrating its outstanding capabilities. At the same time, the commercial model is also significantly ahead of other open source medical LLMs except HuatuoGPT-II. Considering that it has been a long time since most medical LLMs were open sourced, we believe that the new generation of open source medical LLMs will quickly close the gap. After trained on CMB-Exam-Train, DISC-MedLLM ranked second among open source medical models, indicating that the large amount of deterministic medical knowledge contained in multiple-choice questions in CMB-Exam-Train is very helpful for improving performance. How to efficiently inject the knowledge of multiple-choice questions into LLM is a promising task.

\paragraph{On Different Categories.}
LLMs show varied performance across clinical specialties. Specifically, scores for pharmacist-related questions tend to be lower, while those concerning nursing staff are typically higher. This difference might arise from the foundational knowledge nurses require, which is straightforward, compared to the intricate distinctions in drug names and indications pharmacists deal with. Despite these performance variations among specialties, the models exhibit a consistent trend, suggesting no inherent bias towards any particular domain.






\paragraph{On Prompt Strategies}
For the vast majority of domestic General LLM and Medical LLM, both the Few-shot and CoT strategies have little effect on improving model accuracy. The few-shot strategy has improved significantly for models such as Deepseek-llm, Mistral-7B, Mixtral-8x7B, which originally had limited support for Chinese, and smaller models such as Qwen-1.8B and Yi-6B. The CoT strategy even has negative effects on models such as Mistral and ChatGLM-Med, which have very low original accuracy. In CMB-Exam, for problems that do not require reasoning, the CoT strategy may confuses correct information with irrelevant context, thereby reducing accuracy.


\begin{table*}[htbp]
\centering
\vspace{-10pt}
\resizebox{\textwidth}{!}{
\setlength{\tabcolsep}{3pt}
\begin{tabular}{@{}lccccccccc@{}}
\toprule
Aspects& GPT-4 & Yi-34B & Qwen-72B & ChatGPT & Baichuan2-13B & ChatGLM3-6B & Internlm-20B & Deepseekllm-67B & Mixtral-8x7B \\ \midrule
Fluency& 4.95  & 4.99 & 4.96  & 4.97 & 4.93 & 4.92  & 4.9 & 4.78& 2.53\\
Relevance & 4.71  & 4.69& 4.58  & 4.49 & 4.41 & 4.11  & 3.91& 4.04& 2.28\\
Completeness & 4.35  & 4.34& 4.12  & 4.12 & 4.03 & 3.74  & 3.25& 2.62& 1.54\\
Proficiency  & 4.66  & 4.64& 4.55  & 4.53 & 4.36 & 4.23  & 4.14& 4.16& 3.04\\ \midrule
Average& 4.67  & 4.67& 4.55  & 4.53 & 4.43 & 4.25  & 4.05& 3.90 & 2.35\\ \bottomrule
\end{tabular}
}
\caption{Results of CMB-Clin on Automatic Evaluation using GPT-4 for General Models.}
\label{tab:Clin_auto_eval_general}
\end{table*}

\begin{table*}[htbp]
\centering
\resizebox{\textwidth}{!}{
\setlength{\tabcolsep}{3pt}
\begin{tabular}{@{}lccccccccc@{}}
\toprule
Aspects  & HuatuoGPT-II-34B & BianQue-2 & DISC-MedLLM & ChatMed-Consult & MedicalGPT & DISC-MedLLM-Train & DoctorGLM & Bentsao & ChatGLM-Med \\ \midrule
Fluency  & 4.96 & 4.86  & 4.82& 4.88& 4.48   & 4.57  & 4.74  & 3.88& 3.55\\
Relevance& 4.61 & 3.52  & 3.24& 3.08& 2.64   & 2.52  & 2.00  & 2.05& 1.97\\
Completeness & 4.31 & 3.02  & 2.75& 2.67& 2.19   & 1.89  & 1.65  & 1.71& 1.61\\
Proficiency  & 4.53 & 3.60  & 3.51& 3.30& 2.89   & 3.19  & 2.30  & 2.58& 2.37\\ \midrule
Average  & 4.60 & 3.75  & 3.58& 3.48& 3.05   & 3.04  & 2.67  & 2.55& 2.38\\ \bottomrule
\end{tabular}
}
\caption{Results of CMB-Clin on Automatic Evaluation using GPT-4 for Medical Models.}
\label{tab:Clin_auto_eval_medical}
\end{table*}

\paragraph{On the Perceived Difficulty}
As shown in Figure \ref{fig:difficulty}, the professional level continues to improve from bottom to top. Only the Nurse category meets expectations with accuracy decreases from bottom to top. For the Physician, Advanced Professional subcategory have the lowest accuracy and Resident Physician have the second lowest accuracy. After sample analysis, we found that the questions covered in the Resident Physician subcategory involve many uncommon details and knowledge, which increases the probability of hallucinations. For Technologist, it's interesting that the accuracy rate is completely opposite to expectations. We found that there are many questions focus on personnel management and communication in Supervising Technologist subcategory, which is indeed what LLMs are good at. For the Pharmacists, there is no obvious trending. But subcategories involving traditional Chinese medicine always have relative low accuracy, indicating that additional data on traditional Chinese medicine still needs to be supplemented.

\section{Experiments on CMB-Clin}

\subsection{Experimental Setup}
\paragraph{Prompt construction}
Every prompt comprises two components: a description that may  (or may not) encompass conversation history $D_i$, and the question $Q_i$.
To integrate the conversation history into the description, we prepend the appropriate roles to each question and reference.

\paragraph{Expert and Automatic Evaluation}
To prove the validity of our evaluation, we engage three annotators with professional medical knowledge to evaluate on a randomly selected subset of 320 responses of 11 models from different tiers.
Equipped with a reference solution, they score each response across four aspects — Fluency, Relevance, Completeness, and Medical Proficiency — using a grading scale from 1 to 5.
The user interface is shown in Appendix~\ref{app:UI}.
To enhance efficiency and reduce expert evaluation costs, we leverage GPT-4 to assess the responses of all models, adhering to the same guidelines as those used in expert evaluations.
The prompt template for the automatic evaluation is detailed in Appendix~\ref{app:prompt_for_evalution}.

\subsection{Benchmarking Results}
\paragraph{On General LLMs}
As shown in Table \ref{tab:Clin_auto_eval_general}, except for Deepseekllm-67B and Mixtral-8x7B, which have insufficient support for Chinese models, the other General LLMs have shown strong dialogue capabilities based on complex medical records. Taking GPT-4 and ChatGPT as the dividing line, Yi-34B has achieved the same medical dialogue capability as GPT-4. Qwen-72B is weaker than GPT-4 but stronger than ChatGPT, and the remaining models are all weaker than ChatGPT. Compared with their strong performance in CMB-Exam, domestic General LLMs still lag behind OpenAI in CMB-Clin, which is closer to real scenarios. Except for the Yi LLMs, the ability of other domestic LLMs to solve real problems does not match their ability to answer multiple-choice questions, suggesting that they may have been specially strengthened for multiple-choice questions. Such model iteration direction actually deviates from actual needs. During the iteration process, we recommend using both CMB-Exam and CMB-Clin for model capability awareness.

\paragraph{On Medical LLMs}
As shown in Table \ref{tab:Clin_auto_eval_medical}, the overall dialogue ability of Medical LLMs is lower than that of General LLMs. Although the three models of DoctorGLM, Bentsao, and ChatGLM-Med all claim to be optimized for consultation, the actual results show that their conversational capabilities have not been enhanced. It is worth noting that although BianQue-2 performed poorly in CMB-Exam, it performed well in CMB-Clin, indicating that it just lacks the ability to do multiple-choice questions and follow instructions. Although HuatuoGPT-II-34B surpasses GPT-4 in CMB-Exam, it still lags behind GPT-4 and is even lower than its base model Yi-34B in CMB-Clin, indicating that multiple rounds of dialogue data need to be added during its training process. It is noted that performance of DISC-MedLLM trained on CMB-Exam-Train drops significantly on CMB-Clin, indicating the need to add other data or reconstruct multiple-choice questions in the form of dialogues.
To enhance the robustness of our findings, we have included supplementary evaluation results in Appendix~\ref{app:robustness}.

\paragraph{On Different Metrics}
Regarding the Fluency indicator, there is not much difference between General LLMs with most LLM above 4.9, but there are still many Medical LLMs models below 4.5, indicating a lack of basic dialogue capabilities. Relevance, Completeness and Proficiency are all highly differentiated indicators, among which Completeness has the lowest average value, indicating that for medical record consultation scenarios, the completeness of the dialogue and obtaining complete information are the most difficult task.
\begin{figure}[]
\centering
\includegraphics[width=0.85\linewidth]{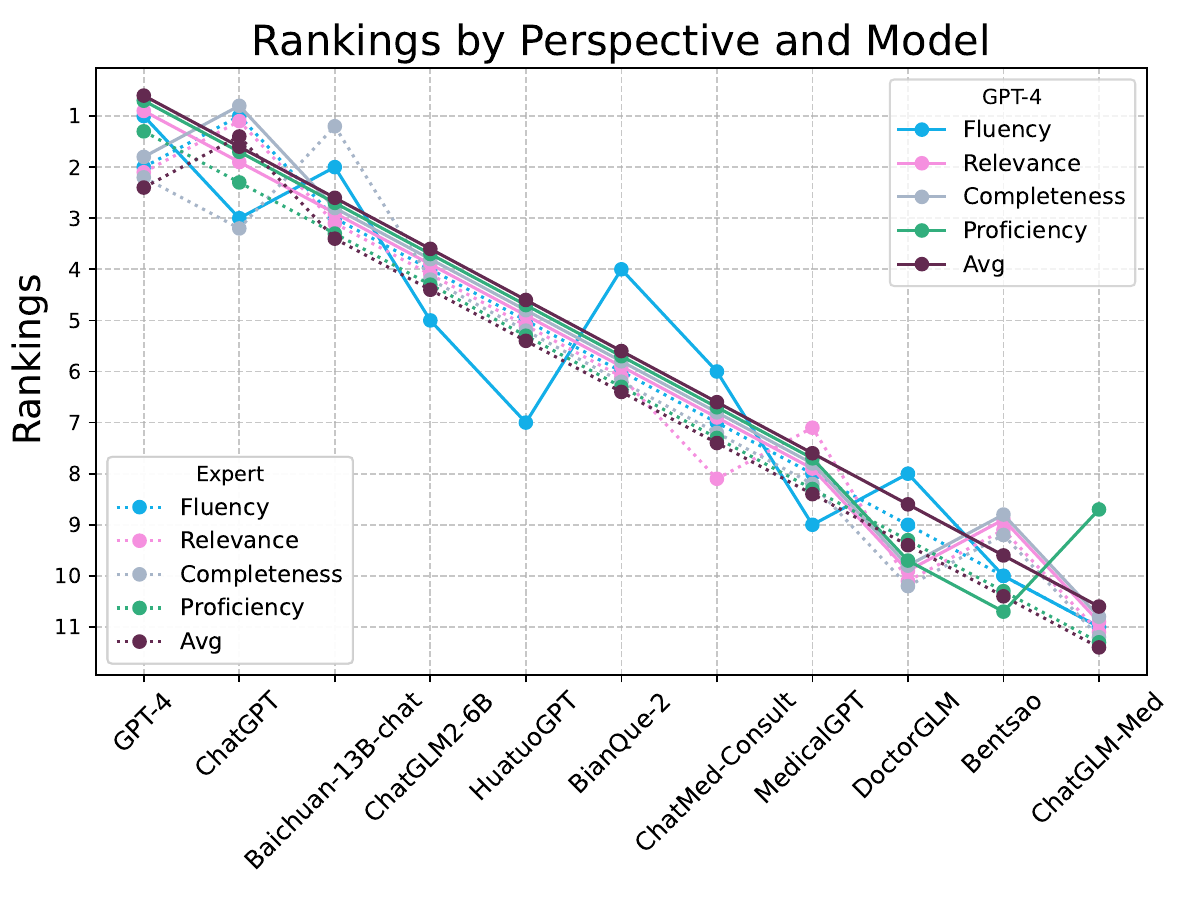}  
\caption{Rankings by perspective and model. Dashed lines and solid lines are the resulted rankings from expert and ChatGPT evaluation, respectively. For visual clarity, each line is shifted vertically for a small value. A model is better if it has a smaller ranking  (a higher position) on the vertical axis.}
\label{fig:QA_ranking_alignment}
\end{figure}

\subsection{Agreements between Automatic and Expert Evaluation}
Figure~\ref{fig:QA_ranking_alignment} demonstrates a strong agreement of resulted rankings between GPT-4 and expert evaluation, with the spearman correlation of rankings being 0.93.
The rankings agree with each other except for a flip for GPT-4 and ChatGPT  (dashed and solid brown lines are parallel, except for a flip at GPT-4 and ChatGPT). 
We also provide a more fine-grained alignment analysis in Appendix~\ref{app:agreement_expert_gpt4}.
The results indicate that the automatic evaluation is highly aligned with expert evaluation.

\section{Conclusion}

Although LLMs have potential in the realm of medicine, their accurate evaluation remains pivotal for real-world applications. The introduction of the CMB benchmark, tailored to the local cultural environment in China, gives a more contextualized and comprehensive evaluation benchmark. Although not framed as a competitive leaderboard, it serves as a crucial tool for tracking LLM progress in medical domains. This might pave the way for a broader and more effective utilization of LLMs in China's medical landscape.

\section*{Ethical Statement}

In terms of Data Anonymity, all data utilized in this study primarily originate from publicly accessible examination questions and coursework exercises that have been processed by experts, and there is no personal information.

In terms of Data Quality, we fully ensure data quality from two aspects: data source and data cleaning. Through strict screening of data sources, the accuracy and authority of the exercises are ensured; through strict data processing, character errors caused by OCR and low-level errors generated during the transcription process are minimized.

In terms of Exaggerating the Abilities of the LLM, we follow the reviewer's suggestions and conducted multiple experiments to ensure the stability of the experimental results. We provide an explanation based on experimental results that provides a possible option for solving the evaluation dilemma of open-ended questions.

\section*{Acknowledgement}
This work was supported by the Shenzhen Science and Technology Program (JCYJ20220818103001002), Shenzhen Doctoral Startup Funding (RCBS20221008093330065), Tianyuan Fund for Mathematics of National Natural Science Foundation of China (NSFC) (12326608).

\section*{Limitations}

The limitations of our study are twofold. 
Firstly, while our benchmark encompasses various subjects in the Chinese medical domain, there remain numerous subjects that necessitate multi-modal capabilities for addressing real-world issues. 
Secondly, within the CMB-Clin section, we standardized the model evaluation method. However, in the real world, diverse medical conditions may require distinct evaluation criteria. Therefore, we advocate the adoption of disease-specific criteria for assessing model performance.

\newpage

\bibliography{anthology,custom}
\bibliographystyle{acl_natbib}

\clearpage

\appendix

\section{Related work}


\subsection{Medical Benchmark}

Medical benchmarks have evolved to broadly encompass two types of tasks based on the capabilities of the models they seek to probe: objective tasks and subjective tasks.
The former typically assumes the form of multiple-choice questions~\citep{MedHop,MedMCQA,MMLU,MultiMedQA}, information retrieval~\citep{LiveQA,HealthQA,MedicationQA}, and cloze-style reading comprehension~\citep{CliCR,emrQA,MASH-QA}, which serve to evaluate a model's medical knowledge with unbiased accuracy. 
Sources for these tasks range from medical textbooks and exams to case reports~\citep{CliCR}, Wikipedia~\citep{MedHop}, and medical practices ~\citep{MMLU,MedMCQA}. 
In contrast, subjective tasks involve open-ended text generation constructed directly from consumer queries and doctor responses, often sourced from online medical forums. 
The task typically demands models to generate consumer-oriented replies~\citep{MultiMedQA,li2023huatuo} or explanations for multiple-choice questions~\citep{CMExam}. 
As of now, there are relatively few open-ended text generation question-answering tasks that specifically center around providing consultation based on diagnostic reports.

Few existing benchmark datasets encapsulate both task types, with PromptCBLUE~\citep{Promptcblue}, MultiMedQA~\citep{MultiMedQA} and CMExam~\citep{CMExam} sharing the closest resemblance to our work. 
Our dataset exceeds in size and includes questions not only from the Chinese National Medical Licensing Examination but also from various authoritative medical textbooks.

Several other datasets have been developed that contribute to the ongoing advancement of medical NLP, albeit with a narrower career focus. For instance, the ExplainCPE~\citep{ExplainCPE} is solely career-specific to pharmacists, while GenMedicalEval~\citep{GenMedeEval}, Medbench~\citep{Medbench} and Medbench\_Opencompass\footnote{\href{https://medbench.opencompass.org.cn/home}{https://medbench.opencompass.org.cn/home}
} are exclusive to physician careers. Our dataset includes questions not only pertaining to physicians but also to nurses, technicians, and pharmacists. These questions are derived from a mix of sources, including the Chinese National Medical Licensing Examination and various authoritative medical textbooks, thereby offering a larger and more comprehensive resource than previously available datasets.


\subsection{Other Benchmarks of Large Language Models}


The explosive growth in the number and capability of LLMs has led to a multitude of works aiming to discern their true capacity, evaluating both their general and specific abilities.  
General ability benchmarks include comprehensive test suites, each targeting different aspects of LLM's proficiency, ranging from handling multi-turn dialogues~\citep{zheng2023judging} to gauging language comprehension and reasoning abilities~\citep{BIG-Bench,zhang2023m3exam,zhong2023agieval}. 

In terms of specific abilities, several benchmarks, apart from those related to medicine, aim to evaluate different capabilities of models. 
ARB~\citep{ARB} was introduced to assess LLMs' performance in high-level reasoning tasks across multiple domains. 
C-Eval~\cite{huang2023c} serves as the first comprehensive benchmark to evaluate the advanced knowledge and reasoning abilities of Chinese-based models. 
Gaokao~\citep{zhang2023evaluating}, MATH~\citep{MATH}, and APPS~\citep{APPS} focus on assessing LLM proficiency in complex, context-specific tasks, and code generation, respectively.




\section{Dataset}
\label{sec:appendix.catalog}
Table \ref{tab:cat-phsician}, \ref{tab:cat-disciplines}, \ref{tab:cat-rest} present a detailed directory structure of CMB-Exam. Initially, the organization is based on clinical professions and the exams commonly undertaken by these professionals, divided into six primary sections. Upon this foundation, each section is further categorized based on career progression and examination subjects. Within each sub-category, we have meticulously classified according to specific departments or courses.

\subsection{Keywords in the TCM subcatalog and Medqa-USMLE}
\label{sec:keywords}
We used jieba\footnote{\url{https://github.com/fxsjy/jieba}} and NLTK\footnote{\url{https://www.nltk.org/}} to perform 2-gram word segmentation on the CMB TCM subcatalog and Medqa-USMLE~\cite{MEDQA} respectively, and compared the top 5 keywords (excluding words with no medical meaning). The results are shown in the Table \ref{tab:key}. It can be seen that the expressions in TCM and English medicine are very different.

\begin{table*}[]
\vspace{-5mm}
\resizebox{1\textwidth}{!}{
\begin{tabular}{@{}l|l@{}}
\toprule
Dataset     & Top 5 Medicine-Related Keywords                                                                              \\ \midrule
CMB-TCM     & 气血 (Qi and blood), 舌苔 (tongue coating), 虚证 (deficiency   syndrome), 病程 (course of disease), 舌淡 (pale tongue) \\
Medqa-USMLE & blood pressure, respiratory rate，physical examination, heart   rate, abdominal pain                          \\ \bottomrule
\end{tabular}
}
\caption{Keywords in the TCM subcatalog and Medqa-USMLE}
\label{tab:key}
\end{table*}

\begin{figure*}[htbp]
\includegraphics[width=\linewidth]{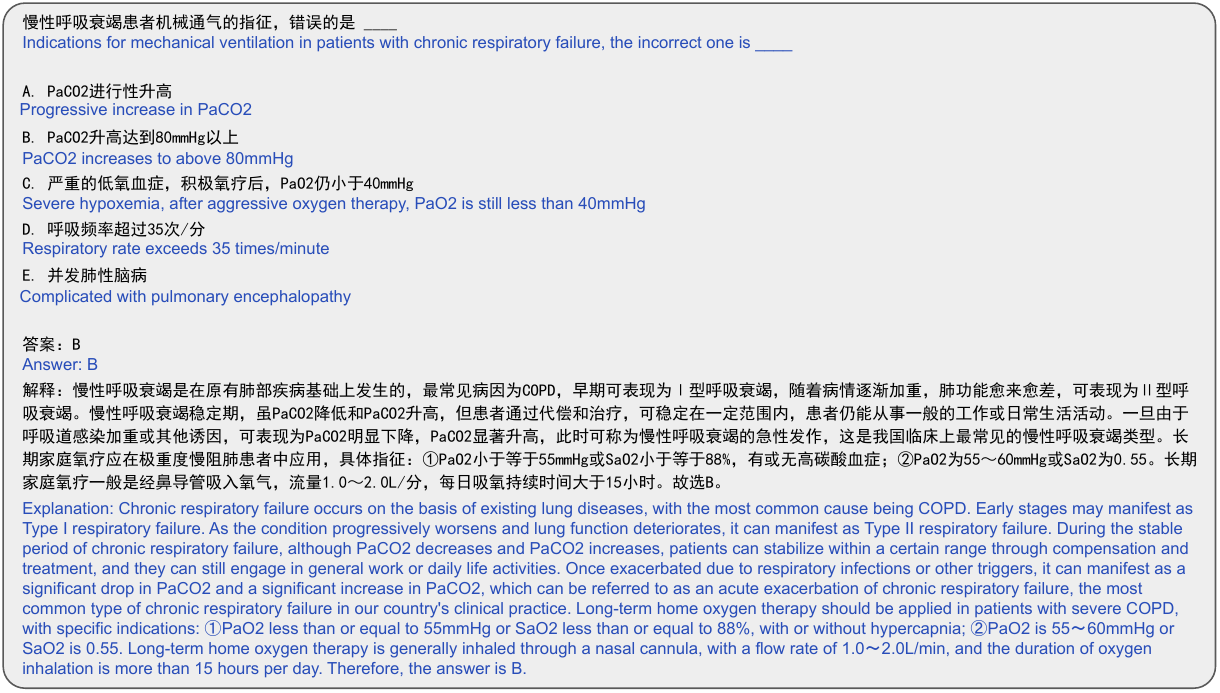}
\centering
\caption{An development example with explanations in CMB-Exam. English translations are shown for better readability.}
\label{fig:CMB-Exam-cot-example}
\end{figure*}

\section{Details of Evaluated Models}
\label{sec:appendix.model}

In this section, we introduce and detail the models utilized in our evaluation. 
These models fall under three primary categories: 12 Chinese medical LLMs, 5 proprietary LLMs, and 13 publicly-available general-domain LLMs.

\textbf{Chinese medical LLMs}:
\begin{itemize}
\item HuatuoGPT-II: HuatuoGPT2 employs an innovative domain adaptation method to significantly boost its medical knowledge and dialogue proficiency. It showcases state-of-the-art performance in several medical benchmarks, especially surpassing GPT-4 in expert evaluations and the fresh medical licensing exams. The number of parameters of the model is 7B, 13B, and 34B.
\item BianQue: It enhances its questioning ability by asking patients for more information to solve the issue that patients may not reveal all information in a single-turn conversation.
\item ChatMed-Consult: It is built upon Chinese LLaMA~\citep{chinese-llama-alpaca} using real-world questions and synthetic responses from ChatGPT.
\item MedicalGPT: It is based on Ziya-LLaMa~\citep{fengshenbang} and adopts a four-stage training recipe, including continued pre-training, supervised fine-tuning, reward modeling, reinforcement learning.
\item ChatGLM-Med: It is finetuned on ChatGLM-6B~\citep{du2022glm} using instruction tuning data, which are built upon CMeKG\footnote{\url{https://github.com/king-yyf/CMeKG_tools}}. 
\item Bentsao: It is finetuned on LLaMa-7B~\citep{touvron2023llama} using the same data as ChatGLM-Med. It's the first Medical LLM trained on LLaMa.
\item DISC-Med: DISC-MedLLM is a large model in the medical field specially designed for medical and health conversational scenarios.
\item DISC-MedLLM-13B (CMB-Exam-Train): The model after fine-tuning DISC-MedLLM on CMB-Exam-Train. ALL of above large language models are fine-tuned for 2 epoch on the full training set with a batch size of 32, with a learning rate of 10−5 using Adam. The warm-up rate of cosine scheduling is set to 0.03.
\item IvyGPT: An LLM based on LLaMA that is trained and fine-tuned with high-quality medical question-answer  (QA) instances and Reinforcement Learning from Human Feedback  (RLHF). 
\item Sunsimiao: Sunsimiao is fine-tuned from Baichuan-7B and ChatGLM-6B series on 100,000-level high-quality Chinese medical data.
\item DoctorGLM: Based on the Chinese consultation model of ChatGLM-6B, it is fine-tuned on a single A100 80G in 13 hours. It's the first Medical LLM trained on ChatGLM.
\end{itemize} 

\textbf{Proprietary models}:
\begin{itemize}
\item ChatGPT: Developed by OpenAI, ChatGPT, rooted in the GPT-3.5 architecture, excels in both understanding and generating natural language.
\item GPT-4: Another offering from OpenAI, GPT-4 employs deep learning techniques to elevate natural language processing capabilities, showcasing remarkable advancements across diverse tasks.
\item JianPeiGPT:A Medical LLM developed by the Pecking Doctor team of Hangzhou Jianpei Technology Co., Ltd \footnote{\url{http://www.jianpeicn.com/}}. The evaluation results were submitted and made public on December 15, 2023 after the opening of CMB.
\item ShukunGPT: A Medical LLM developed by Shukun Technology\footnote{\url{https://www.shukun.net/}}. The evaluation results were submitted and made public on October 23, 2023 after the opening of CMB.
\item AntGLM-Med: A Medical LLM developed by the algorithm research group from AntGroup\footnote{\url{https://www.antgroup.com/en}}. The evaluation results were submitted and made public on December 23, 2023 after the opening of CMB.
\end{itemize}

\textbf{Publicly-available general-domain LLMs}:
\begin{itemize}
\item ChatGLM-3: The third version of ChatGLM, which is an open source, bilingual dialogue language model.
\item Baichuan2-chat: An advanced variant of Baichuan-13B model, focuses on dialogue tasks, boasting 13 billion parameters for efficient and effective conversation generation. The number of parameters of the model is 7B and 13B.
\item Qwen-Chat: Tongyi Qianwen large model series developed by Alibaba Cloud based on Transformer, which is trained on extremely large-scale pre-training data. The number of parameters of the model is 1.8B, 7B and 72B.
\item Yi: Large language models trained from scratch by developers at 01.AI\footnote{https://www.lingyiwanwu.com/}. The number of parameters of the model is 6B and 34B.
\item Deepseek-llm-67B-Chat: An advanced language model comprising 67 billion parameters. It has been trained from scratch on a vast dataset of 2 trillion tokens in both English and Chinese.
\item Internlm-Chat: It's pre-trained on over 2.3T Tokens containing high-quality English, Chinese, and code data. Chat version has undergone SFT and RLHF training, enabling it to better and more securely meet users' needs. The number of parameters of the model is 7B and 20B.
\item Mistral\footnote{\url{https://mistral.ai/}}: A 7B dense Transformer, fast-deployed and easily customisable. Small, yet powerful for a variety of use cases. Supports English and code, and a 8k context window.
\item Mixtral-8x7B-32kseqlen: A 7B sparse Mixture-of-Experts model with stronger capabilities than Mistral 7B. Uses 12B active parameters out of 45B total. Supports multiple languages, code and 32k context window.
\end{itemize}

\section{Experiment Details of CMB-Clin}

\subsection{Screenshot of Human Evaluation UI}
\label{app:UI}
\begin{figure*}[]
\includegraphics[width=0.75\linewidth]{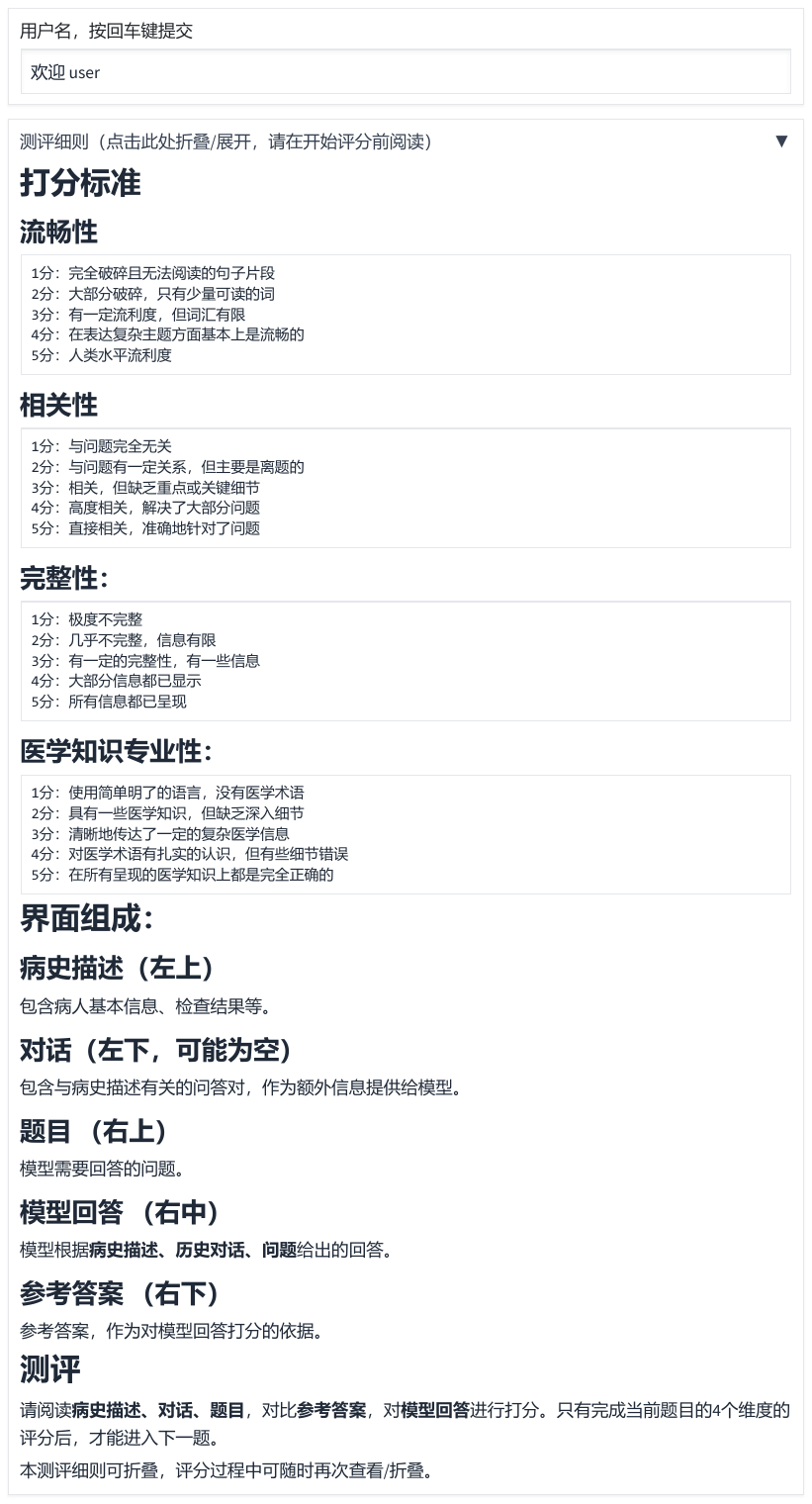}
\centering
\caption{The guideline for human evaluation and the introduction to components of user interface  (in Chinese). Note that Figure~\ref{fig:QA_human_eval_UI_1} precedes Figure~\ref{fig:QA_human_eval_UI_2} in the same webpage.}
\label{fig:QA_human_eval_UI_1}
\end{figure*}

\begin{figure*}[]
\includegraphics[width=1.\linewidth]{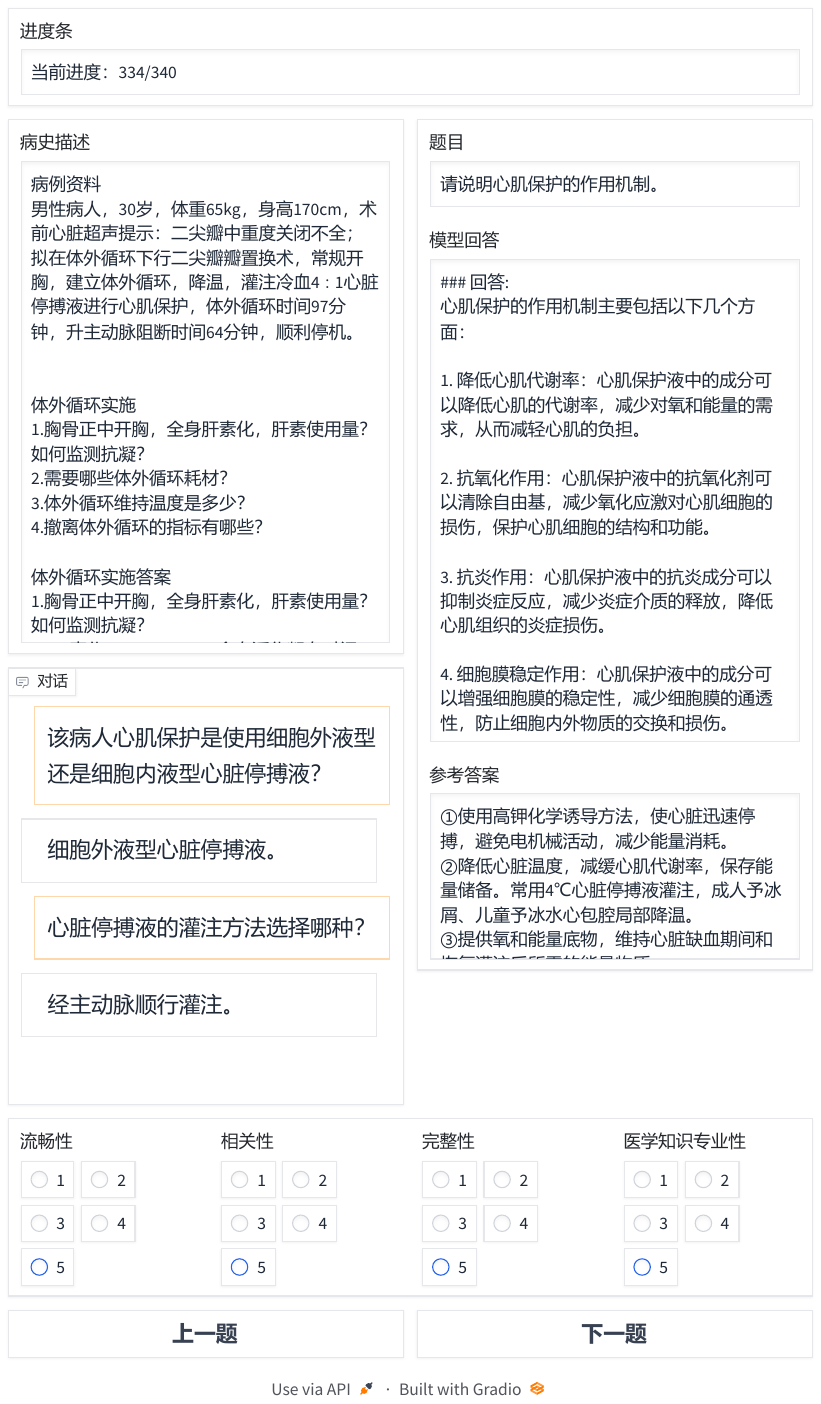}
\caption{The user interface for scoring an answer  (in Chinese). Note that Figure~\ref{fig:QA_human_eval_UI_2} follows Figure~\ref{fig:QA_human_eval_UI_1} in the same webpage.}
\label{fig:QA_human_eval_UI_2}
\end{figure*}

We show the screenshot of human evaluation UI in Figure~\ref{fig:QA_human_eval_UI_1} and Figure~\ref{fig:QA_human_eval_UI_2}. We split the screenshot into two figures for better visual clarity.

\subsection{Prompts for Automatic Evaluation}
\label{app:prompt_for_evalution}
The prompt for automatic evaluation contains task instructions, metrics, criteria, and placeholders for information to be evaluated. 
It is designed based on the suggestion of experts and used by both ChatGPT and GPT-4.
\lstset{
 backgroundcolor=\color[RGB]{245,245,244},
 breaklines=true,
 basicstyle=\ttfamily\small
}
\begin{lstlisting}
You are an AI evaluator specializing in assessing the quality of answers provided by other language models. Your primary goal is to rate the answers based on their fluency, relevance, completeness, proficiency in medicine. Use the following scales to evaluate each criterion:
Fluency:
1: Completely broken and unreadable sentence pieces
2: Mostly broken with few readable tokens
3: Moderately fluent but with limited vocabulary
4: Mostly coherent in expressing complex subjects
5: Human-level fluency

Relevance:
1: Completely unrelated to the question
2: Some relation to the question, but mostly off-topic
3: Relevant, but lacking focus or key details
4: Highly relevant, addressing the main aspects of the question
5: Directly relevant and precisely targeted to the question

Completeness:
1: Extremely incomplete
2: Almost incomplete with limited information
3: Moderate completeness with some information
4: Mostly complete with most of the information displayed
5: Fully complete with all information presented

Proficiency in medicine:
1: Using plain languages with no medical terminology.
2: Equipped with some medical knowledge but lacking in-depth details
3: Conveying moderately complex medical information with clarity
4: Showing solid grasp of medical terminology but having some minor mistakes in detail 
5: Fully correct in all presented medical knowledge

You will be provided with the following information:
- a description
- a conversation based on the description  (optional)
- a question based on the description and conversation
- the solution to the question
- a model's answer to the question

[description]
{description}
[end of description]

[conversation]
{history}
[end of conversation]

[question]
{question}
[end of question]

[solution]
{solution}
[end of solution]

[answer]
{answer}
[end of answer]
Make sure to provide your evaluation results in JSON format and ONLY the JSON, with separate ratings for each of the mentioned criteria as in the following example:
{`fluency': 3, `relevance': 3, `completeness': 3, `proficiency': 3} 
\end{lstlisting}

\begin{table*}[]
\vspace{-5mm}
\centering
\small
\begin{tabular}{@{}lccccccccc@{}}
\toprule
\textbf{Models} & \textbf{Fluency} & \textbf{Relevance} & \textbf{Completeness} & \textbf{Proficiency} & \textbf{Avg.} \\ 
\midrule
ChatGPT  & \textbf{4.93} & \textbf{4.65}& \textbf{4.22}& 4.34  & \textbf{4.53} \\
GPT-4  & 4.88 & 4.61& 4.20& \textbf{4.39}  & 4.52 \\
Baichuan-13B-chat & 4.79 & 4.29& \textbf{4.22}& 4.30  & 4.40 \\
ChatGLM2-6B & 4.77 & 4.06& 3.96& 3.99  & 4.20 \\
HuatuoGPT& 4.70 & 3.89& 3.69& 3.81  & 4.02 \\
BianQue-2& 4.44 & 3.50& 3.30& 3.43  & 3.67 \\
ChatMed-Consult& 4.26 & 3.39& 3.16& 3.27  & 3.52 \\
MedicalGPT  & 4.21 & 3.40& 3.09& 3.10  & 3.45 \\
DoctorGLM& 3.74 & 2.46& 2.35& 2.30  & 2.71 \\
Bentsao  & 3.52 & 2.62& 2.36& 2.30  & 2.70 \\
ChatGLM-Med & 2.92 & 2.23& 1.98& 1.92  & 2.26 \\
\bottomrule
\end{tabular}
\caption{Results of \textit{expert} evaluation on CMB-Clin. \textit{Avg.} are the averaged scores of each model over all perspectives.  Models are arranged in descending order of \textit{Avg.}}
\label{tab:Clin_expert_eval}
\end{table*}

\begin{table}[htbp]\footnotesize
 \begin{tabular}{l*{5}{c}}
  \toprule
  Settings & Original & T-0.2 & T-0.6 & T-1.0 & T-1.5 \\
  \midrule
  Original & 1.00 & 0.95 & 0.90 & 0.87 & 0.87 \\
  T-0.2 & 0.95 & 1.00 & 0.98 & 0.88 & 0.88 \\
  T-0.6 & 0.90 & 0.98 & 1.00 & 0.90 & 0.90 \\
  T-1.0 & 0.87 & 0.88 & 0.90 & 1.00 & 1.00 \\
  T-1.5 & 0.87 & 0.88 & 0.90 & 1.00 & 1.00 \\
  \bottomrule
 \end{tabular}
 \caption{Pairwise Spearman correlations between results under different decoding temperatures. Original: results of greedy decoding  (temperature 0). T-$x$: results of using nucleus sampling under temperature $x$.}
 \label{tab:qa_decoding_params}
\end{table}

\begin{table*}[]
\centering
\vspace{-8pt}
\resizebox{\textwidth}{!}{
\setlength{\tabcolsep}{3pt}
\begin{tabular}{@{}lccccccccc@{}}
\toprule
Aspects & GPT-4 & Yi-34B & Qwen-72B & ChatGPT & Baichuan2-13B & ChatGLM3-6B & Internlm-20B & Deepseekllm-67B & Mixtral-8x7B \\ \midrule
Fluency & 4.94 $\pm$ 0.06 & 5.00 $\pm$ 0.00 & 5.00 $\pm$ 0.00 & 5.00 $\pm$ 0.00 & 4.96 $\pm$ 0.01 & 4.96 $\pm$ 0.01 & 4.92 $\pm$ 0.01 & 4.82 $\pm$ 0.07 & 2.38 $\pm$ 0.20 \\
Relevance & 4.54 $\pm$ 0.17 & 4.77 $\pm$ 0.04 & 4.51 $\pm$ 0.05 & 4.43 $\pm$ 0.06 & 4.13 $\pm$ 0.05 & 4.07 $\pm$ 0.05 & 3.98 $\pm$ 0.08 & 4.12 $\pm$ 0.08 & 2.25 $\pm$ 0.13 \\
Completeness & 4.10 $\pm$ 0.24 & 4.65 $\pm$ 0.04 & 4.06 $\pm$ 0.06 & 4.15 $\pm$ 0.07 & 3.85 $\pm$ 0.04 & 3.74 $\pm$ 0.07 & 3.48 $\pm$ 0.06 & 2.73 $\pm$ 0.07 & 1.56 $\pm$ 0.08 \\
Proficiency & 4.45 $\pm$ 0.29 & 4.76 $\pm$ 0.06 & 4.47 $\pm$ 0.08 & 4.45 $\pm$ 0.07 & 4.17 $\pm$ 0.05 & 4.12 $\pm$ 0.05 & 4.15 $\pm$ 0.07 & 4.18 $\pm$ 0.11 & 2.73 $\pm$ 0.16 \\ \midrule
Average & 4.51 & 4.79 & 4.51 & 4.51 & 4.28 & 4.22 & 4.13 & 3.96 & 2.23 \\\bottomrule
\end{tabular}
}
\caption{Results of multiple runs on CMB-Clin using GPT-4 for \textbf{general models}. Each cell in the first four rows show the mean and standard deviation of repeating the experiments 3 times. The last row shows the average values of means in their corresponding column.}
\label{tab:Clin_repeated_general}
\end{table*}

\begin{table*}[]
\centering
\vspace{-5pt}
\resizebox{\textwidth}{!}{
\setlength{\tabcolsep}{3pt}
\begin{tabular}{@{}lccccccccc@{}}
\toprule
 & HuatuoGPT-II-34B & BianQue-2 & DISC-MedLLM & ChatMed-Consult & MedicalGPT & DISC-MedLLM-Train & DoctorGLM & Bentsao & ChatGLM-Med \\ \midrule
Fluency & 5.00 $\pm$ 0.00 & 4.97 $\pm$ 0.01 & 5.00 $\pm$ 0.00 & 4.97 $\pm$ 0.01 & 4.54 $\pm$ 0.02 & 4.48 $\pm$ 0.07 & 4.89 $\pm$ 0.03 & 4.01 $\pm$ 0.05 & 3.63 $\pm$ 0.02 \\
Relevance & 4.55 $\pm$ 0.05 & 3.55 $\pm$ 0.07 & 3.25 $\pm$ 0.07 & 3.05 $\pm$ 0.06 & 2.62 $\pm$ 0.04 & 2.52 $\pm$ 0.07 & 1.97 $\pm$ 0.07 & 2.06 $\pm$ 0.06 & 2.01 $\pm$ 0.04 \\
Completeness & 4.36 $\pm$ 0.05 & 3.07 $\pm$ 0.06 & 2.71 $\pm$ 0.07 & 2.62 $\pm$ 0.07 & 2.22 $\pm$ 0.03 & 1.95 $\pm$ 0.05 & 1.71 $\pm$ 0.05 & 1.66 $\pm$ 0.03 & 1.64 $\pm$ 0.02 \\
Proficiency & 4.48 $\pm$ 0.04 & 3.71 $\pm$ 0.07 & 3.52 $\pm$ 0.12 & 3.38 $\pm$ 0.1 & 2.89 $\pm$ 0.1 & 3.07 $\pm$ 0.1 & 2.31 $\pm$ 0.11 & 2.58 $\pm$ 0.14 & 2.42 $\pm$ 0.06 \\ \midrule
Average & 4.60 & 3.82 & 3.62 & 3.51 & 3.07 & 3.01 & 2.72 & 2.58 & 2.43 \\\bottomrule
\end{tabular}
}
\caption{Results of multiple runs on CMB-Clin using GPT-4 for \textbf{medical models}. Each cell in the first four rows show the mean and standard deviation of repeating the experiments 3 times. The last row shows the average values of means in their corresponding column.}
\label{tab:Clin_repeated_med}
\end{table*}

\begin{table*}[]\scriptsize
\centering
\setlength{\tabcolsep}{9pt}
\renewcommand{\arraystretch}{0.7} 
\begin{tabular}{@{}c|c|c|c@{}}
\toprule
Category& Subcategory& Subject& \# Questions\\ \midrule
\multirow{81}{*}{Physician} & \multirow{18}{*}{Resident Physician}& Clinical Pathology & 1124  \\
&& Oral& 1074  \\
&& Otolaryngology & 952\\
&& Rehabilitation Medicine& 461\\
&& Ophthalmology  & 951\\
&& Neurology  & 791\\
&& Orthopedics& 939\\
&& Anesthesiology & 907\\
&& Pediatrics & 749\\
&& Dermatology& 977\\
&& Psychiatry & 903\\
&& General Practice& 712\\
&& Medical Imaging& 964\\
&& Internal Medicine  & 752\\
&& Ultrasound & 430\\
&& Surgery& 829\\
&& Obstetrics and Gynecology  & 800\\
&& Pediatric Surgery  & 296\\ \cmidrule (l){2-4} 
& \multirow{5}{*}{Licensed Assistant Physician}  & Integrated Chinese and Western Medicine  & 3441  \\
&& Clinical& 5364  \\
&& Chinese Medicine& 3454  \\
&& Public Health  & 2067  \\
&& Oral& 1090  \\ \cmidrule (l){2-4} 
& \multirow{5}{*}{Licensed Physician}& Chinese Medicine& 4490  \\
&& Public Health  & 4085  \\
&& Clinical& 10241 \\
&& Oral& 1505  \\
&& Integrated Chinese and Western Medicine& 5320  \\ \cmidrule (l){2-4} 
& \multirow{43}{*}{Associate Professional Physician} & General Medicine& 3492  \\
&& Internal Oral  & 858\\
&& Orthopedics& 894\\
&& Chinese Internal Medicine  & 2896  \\
&& Surgery& 5071  \\
&& Ultrasound Medicine& 2218  \\
&& Dermatology and Venereology& 1158  \\
&& Otolaryngology & 983\\
&& Internal Medicine  & 5671  \\
&& Infectious Diseases& 600\\
&& Obstetrics and Gynecology  & 2641  \\
&& Cardiovascular Internal Medicine and Respiratory Internal Medicine & 617\\
&& Oncology& 942\\
&& Acupuncture Attending in TCM& 1169  \\
&& Pathology  & 1642  \\
&& Preventive Medicine& 2817  \\
&& Pediatrics & 3773  \\
&& Psychotherapy  & 1393  \\
&& Radiology  & 2401  \\
&& Psychiatry & 754\\
&& Oral Restoration& 1183  \\
&& Dermatology& 909\\
&& Digestive Internal Medicine& 160\\
&& Rehabilitation Medicine& 630\\
&& Infectious Disease & 861\\
&& Nuclear Medicine& 1250  \\
&& Oral Medicine  & 862\\
&& Integrated Chinese and Western Internal Medicine& 1101  \\
&& Ophthalmology  & 988\\
&& Anesthesiology & 923\\
&& Hospital Infection & 827\\
&& Nutrition  & 1009  \\
&& Tuberculosis& 58\\
&& Critical Care Medicine & 579\\
&& Psychological Counselor& 495\\
&& Pain Medicine  & 884\\
&& Neurology  & 126\\
&& Orthodontics& 578\\
&& Oral and Maxillofacial Surgery & 367\\
&& Plastic Surgery& 187\\
&& Nephrology & 81\\
&& Rheumatology and Clinical Immunology& 37\\
&& Occupational Disease& 54\\ \cmidrule (l){2-4} 
& \multirow{10}{*}{Advanced Professional Physicians} & Respiratory InternalMedicine& 1522  \\
&& Orthopedics& 1245  \\
&& Endocrinology  & 1326  \\
&& Cardiology & 1604  \\
&& Digestive Internal Medicine& 1577  \\
&& General Surgery Senior & 1850  \\
&& Gynecology and Obstetrics  & 3249  \\
&& General Internal Medicine  & 607\\
&& General Practice& 74\\
&& Pediatrics & 65\\ \bottomrule
\end{tabular}
\caption{Catalog Structure of Physician}
\label{tab:cat-phsician}
\end{table*}

\begin{table*}[h]\scriptsize
\centering
\setlength{\tabcolsep}{9pt}
\renewcommand{\arraystretch}{1} 
\begin{tabular}{@{}c|c|c|c@{}}
\toprule
Category & Subcategory& Subject&\# Questions  \\ \midrule
\multirow{53}{*}{Undergraduate Disciplines} & \multirow{17}{*}{Foudamental Medicine}  & Pathophysiology& 1455 \\
  & & Medical Psychology& 932  \\
  & & Biochemistry and MolecularBiology  & 2402 \\
  & & Cell Biology  & 1399 \\
  & & Medical Immunology& 2485 \\
  & & Pathology & 2786 \\
  & & Medical Genetics  & 1369 \\
  & & Parasitology  & 806  \\
  & & Systematic Anatomy& 1967 \\
  & & Bioinformatics& 185  \\
  & & Physiology& 2306 \\
  & & Pharmacology  & 2424 \\
  & & Medical Microbiology  & 1342 \\
  & & Local Anatomy & 489  \\
  & & Histology and Embryology  & 1398 \\
  & & Human Parasitology& 766  \\
  & & Medical Statistics& 198  \\ \cmidrule (l){2-4} 
  & \multirow{22}{*}{Clinical Medicine} & Medical Imaging& 1858 \\
  & & Radiology & 541  \\
  & & Experimental Diagnostic Medicine  & 548  \\
  & & Neurology & 1163 \\
  & & Surgery& 2164 \\
  & & Dermatology and Venereology& 2168 \\
  & & Pediatrics& 3760 \\
  & & Nuclear Medicine  & 1383 \\
  & & Physical Diagnosis& 621  \\
  & & Dental Pulp Disease& 346  \\
  & & Basic Nursing & 978  \\
  & & Diagnostics& 103  \\
  & & Ultrasonic Medicine& 192  \\
  & & Oral Care & 263  \\
  & & Evidence-Based Medicine& 95\\
  & & Fundamental Nursing& 393  \\
  & & Epidemiology  & 864  \\
  & & Oral Tissue Pathology & 387  \\
  & & Infectious Disease& 287  \\
  & & Oral Anatomy and Physiology& 362  \\
  & & Anesthesiology& 606  \\
  & & Interventional Radiology  & 81\\ \cmidrule (l){2-4} 
  & \multirow{3}{*}{TCM and Chinese Herbal Medicine}& Preventive Medicine& 1926 \\
  & & Hygiene& 1316 \\
  & & Medical Ethics& 500  \\ \cmidrule (l){2-4} 
  & \multirow{11}{*}{Preventive Medicine and Public Health} & TCM Ophthalmology & 915  \\
  & & Essential Prescriptions Worth a Thousand Gold & 1051 \\
  & & Basic Theories of TCM & 2706 \\
  & & TCM Diagnosis & 2036 \\
  & & TCM& 1921 \\
  & & Warm Disease Theory& 1088 \\
  & & History of Chinese Medicine& 662  \\
  & & TCM Internal Medicine & 1738 \\
  & & TCM Pediatrics& 694  \\
  & & Treatise on Cold Pathogenic Diseases  & 1390 \\
  & & Lecture on Inner Canon& 456  \\ \bottomrule
\end{tabular}
\caption{Catalog Structure of Undergraduate Disciplines}
\label{tab:cat-disciplines}
\end{table*}

\begin{table*}[h]\scriptsize
\centering
\setlength{\tabcolsep}{17pt}
\renewcommand{\arraystretch}{1} 
\begin{tabular}{@{}c|c|c|c@{}}
Category  & Subcategory& Subject& \# Questions  \\ \midrule
\multirow{8}{*}{Nurse} & Practicing Nurse& Practicing Nurse  & 3303 \\ \cmidrule (l){2-4} 
& Licensed Practical Nurse& Licensed Practical Nurse  & 4223 \\ \cmidrule (l){2-4} 
& \multirow{5}{*}{Charge Nurse}  & Pediatric & 905  \\
&& Internal Medicine & 958  \\
&& Charge Nurse  & 4558 \\
&& Surgery& 341  \\
&& Obstetrics and Gynecology & 755  \\ \cmidrule (l){2-4} 
& Advanced Practice Nurse& Advanced Practice Nurse& 1876 \\ \midrule
\multirow{21}{*}{Technician}  & \multirow{4}{*}{Medical Technician}& Rehabilitation Medicine Therapy& 1752 \\
&& Radiology & 1033 \\
&& Inspection& 1166 \\
&& Oncology  & 1086 \\ \cmidrule (l){2-4} 
& \multirow{4}{*}{Medical Technologist}  & Rehabilitation Medicine Therapy& 1739 \\
&& Oncology  & 1538 \\
&& Radiology & 1337 \\
&& Inspection& 1458 \\ \cmidrule (l){2-4} 
& \multirow{13}{*}{Supervising Technologist} & Radiation Therapy for Oncology& 1701 \\
&& Ultrasonic Medicine& 145  \\
&& Blood Transfusion Technology  & 2199 \\
&& Microbiological Inspection& 704  \\
&& Radiology & 1428 \\
&& Pathology & 2407 \\
&& Physical and Chemical Inspection  & 783  \\
&& Clinical Medicine Inspection  & 1378 \\
&& Medical Record Information& 1331 \\
&& Nuclear Medicine  & 1275 \\
&& Electrocardiology & 1021 \\
&& Disinfection Technology& 575  \\
&& Rehabilitation Medicine and Treatment & 948  \\ \midrule
\multirow{5}{*}{Graduate Entrance Exam} & \multirow{2}{*}{Nursing}& Surgical Nursing  & 1112 \\
&& Basic Nursing & 902  \\ \cmidrule (l){2-4} 
& Political Science  & Political Science & 1514 \\ \cmidrule (l){2-4} 
& Integrated Western Medicine& Integrated Western Medicine& 8913 \\ \cmidrule (l){2-4} 
& Integrated TCM & Integrated TCM& 3924 \\ \midrule
\multirow{8}{*}{Pharmacist}& Licensed Pharmacist& Licensed Pharmacist  & 8248 \\ \cmidrule (l){2-4} 
& Licensed TCM Pharmacist& Licensed TCM Pharmacist& 4460 \\ \cmidrule (l){2-4} 
& Junior Pharmacist  & Junior Pharmacist & 2720 \\ \cmidrule (l){2-4} 
& Junior Pharmacist Assistant& Junior Pharmacist Assistant& 3705 \\ \cmidrule (l){2-4} 
& Junior TCM Pharmacist  & Junior TCM Pharmacist Assistant& 3502 \\ \cmidrule (l){2-4} 
& Junior TCM Pharmacist  & Junior TCM Pharmacist Assistant& 4017 \\ \cmidrule (l){2-4} 
& Chief Pharmacist& Chief Pharmacist  & 3403 \\ \cmidrule (l){2-4} 
& Chief TCM Pharmacist& Chief TCM Pharmacist  & 3299 \\ \bottomrule
\end{tabular}
\caption{Catalog Structure of Nurse, Technician, Graduate Entrance Exam and Pharmacist}

\label{tab:cat-rest}
\end{table*}

\begin{figure*}[h]
\centering

\begin{minipage}{\linewidth}
\centering
\setlength{\tabcolsep}{4.6pt}
\renewcommand{\arraystretch}{0.7} 
\scriptsize

\end{minipage}

\vspace{20pt} 


\vspace{20pt} 

\begin{minipage}{\linewidth}
\vspace{-8mm}
\includegraphics[width=1.\linewidth]{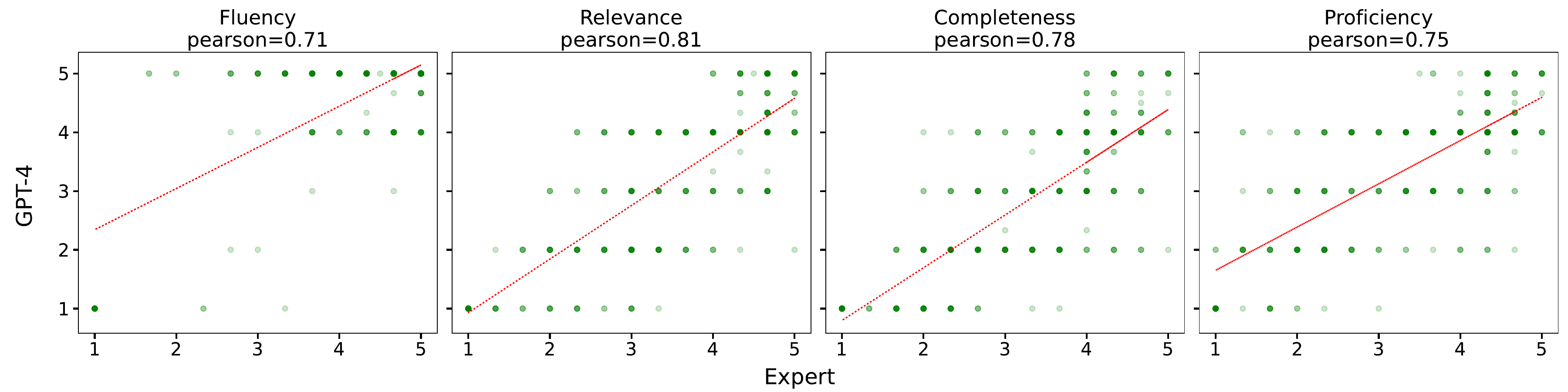}
\caption{Correlation of expert and automatic evaluation on CMB-Clin of each perspective with pearson correlation. The four plots show correlations in fluency, relevance, completeness and proficiency in medicine, respectively. Each plot consists of 320 data points with many overlapped. The darker a point is, the more overlapped data there are at that position. Each expert score is averaged over the three expert annotators.}
\vspace{-3mm}
\label{fig:QA_alignment_perspective}
\end{minipage}
\end{figure*}



\section{Additional Results on CMB-Clin}
\subsection{Results of Expert evaluation}
320 model responses are randomly sampled for this experiment due to a large number of answers to be evaluated and limited expert resources. We present the detailed results of expert evaluation in Table~\ref{tab:Clin_expert_eval}.

\label{app:detailed_scores_clin}

\subsection{Robustness Experiments on CMB-Clin}\label{app:robustness}
To show the robustness of GPT-4-as-a-judge, we also present the results of 3 independent runs for each  general and medical LLMs in Table~\ref{tab:Clin_repeated_general} and Table~\ref{tab:Clin_repeated_med}, respectively. All experiments are conducted between January 20th and January 26th, 2024. 

For Medical LLMs, as shown in Table~\ref{tab:Clin_repeated_med}, although the absolute scores deviate from the results in Table~\ref{tab:Clin_auto_eval_medical}, their rankings remain the same. 
For General LLMs, as shown in Table~\ref{tab:Clin_repeated_general}, the rankings remain the same except for Yi-34B. The standard deviations shown in each cell indicate the robustness of evaluating CMB-Clin with GPT-4. 


\subsection{Agreement of Expert and GPT-4 Evaluation}\label{app:agreement_expert_gpt4}
Figure~\ref{fig:QA_alignment_perspective} shows the agreement between expert and GPT-4 evaluation on each perspective. The pearson correlations are all above 0.71, indicating a strong linear correlation between the two evaluation approaches.


\subsection{Pairwise Correlation of Rankings under Different Temperatures}
We evaluate the results generated under each setting  (\textit{i.e.,} under different temperatures) using ChatGPT. Then for each setting, we obtain a ranking for all models. We then calculate the pairwise spearman correlation between all sets of rankings. The results are summarized in Table~\ref{tab:qa_decoding_params}.






\end{CJK}
\end{document}